\documentclass{article}

\usepackage{arxiv}
\usepackage{blindtext}
\usepackage{csquotes}
\usepackage{enumitem,amssymb}
\usepackage{bm}
\usepackage{pifont}
\usepackage{amsmath}
\usepackage{graphicx}
\usepackage{subfigure}
\usepackage{placeins}
\usepackage{algpseudocode}
\usepackage{algorithm}
\usepackage{rotating}
\usepackage{booktabs}
\usepackage{subcaption}
\usepackage{tikz}
\usepackage{adjustbox}
\usepackage{booktabs}
\usepackage{enumitem}
\usepackage[format=plain]{caption}
\usepackage{tcolorbox}
\usepackage{lipsum}
\tcbuselibrary{skins,breakable}
\usetikzlibrary{shadings,shadows}
\usepackage[parfill]{parskip}
\usepackage{stackengine}
\usepackage{amsthm}
\usepackage[T1]{fontenc}    % use 8-bit T1 fonts
\usepackage{url}            % simple URL typesetting
\usepackage{amsfonts}       % blackboard math symbols
\usepackage{nicefrac}       % compact symbols for 1/2, etc.
\usepackage{microtype}      % microtypography
\usepackage[numbers]{natbib}
\usepackage[hidelinks]{hyperref}       % hyperlinks
\usepackage{lastpage}
{\endtcolorbox}
\providecommand{\acknowledgement}[1]{\textit{\textbf{Acknowledgements} #1}}

\usetikzlibrary{bayesnet}
\usetikzlibrary{arrows}
\usetikzlibrary{shapes,arrows,calc}
\usetikzlibrary {shapes.geometric}

\pgfdeclarelayer{background}
\pgfdeclarelayer{foreground}
\pgfsetlayers{background,main,foreground}

\newlist{todolist}{itemize}{2}
\setlist[todolist]{label=$\square$}

\newcommand\indep{\protect\mathpalette{\protect\independenT}{\perp}}
\def\independenT#1#2{\mathrel{\rlap{$#1#2$}\mkern2mu{#1#2}}}

\definecolor{ibmorange}{HTML}{FE6100}
\definecolor{ibmpurple}{HTML}{785EF0}
\definecolor{ibmblue}{HTML}{648FFF}
\definecolor{ibmpink}{HTML}{DC267F}
\definecolor{tolsand}{HTML}{DDCC77}
\definecolor{ibmaqua}{HTML}{31C9B0}

\newcommand{\mublend}{\mu_{\textup{blend}}}
\newcommand{\sigmablend}{\sigma_{\textup{blend}}}

\newtheorem{examples}{Example}
\newtheorem{remarks}{Remark}
\newtheorem{theorems}{Theorem}
\newtheorem{lemmas}{Lemma}
\newtheorem{propositions}{Proposition}
\newtheorem{corollaries}{Corollary}
\newtheorem*{proofs}{Proof}

\title{CoCoAFusE: Beyond Mixtures of Experts\\via Model Fusion}
\author{%
    {Aurelio~Raffa Ugolini}\\
    {Dip. di Elettronica, Informazione e Bioingegneria}\\
    {Politecnico di Milano, 20133 (Italy)}\\
    {\texttt{aurelio.raffa@polimi.it}}
    \And
    {Mara Tanelli}\\
    {Dip. di Elettronica, Informazione e Bioingegneria}\\
    {Politecnico di Milano, 20133 (Italy)}\\
    {\texttt{mara.tanelli@polimi.it}}
    \And
    {Valentina Breschi}\\
    {Dept. of Electrical Engineering}\\
    {Eindhoven University of Technology, 5600 MB (The Netherlands)}\\
    {\texttt{v.breschi@tue.nl}}
}

\begin{document}
	\maketitle

	\begin{abstract}%
		Many learning problems involve multiple patterns and varying degrees of uncertainty dependent on the covariates. Advances in Deep Learning (DL) have addressed these issues by learning highly nonlinear input-output dependencies. However, model interpretability and Uncertainty Quantification (UQ) have often straggled behind. In this context, we introduce the Competitive/Collaborative Fusion of Experts (CoCoAFusE), a novel, Bayesian Covariates-Dependent Modeling technique.
		CoCoAFusE builds on the very philosophy behind Mixtures of Experts (MoEs), blending predictions from several simple sub-models (or ``experts'') to achieve high levels of expressiveness while retaining a substantial degree of local interpretability. Our formulation extends that of a classical Mixture of Experts by contemplating the \textit{fusion} of the experts' distributions in addition to their more usual mixing (i.e., superimposition). Through this additional feature, CoCoAFusE better accommodates different scenarios for the intermediate behavior between generating mechanisms, resulting in tighter credible bounds on the response variable. Indeed, only resorting to mixing, as in classical MoEs, may lead to multimodality artifacts, especially over smooth transitions.
		Instead,
		CoCoAFusE can avoid these artifacts even under the same structure and priors for the experts, leading to greater expressiveness and flexibility in modeling.
		This new approach is showcased extensively on a suite of motivating numerical examples and a collection of real-data ones, demonstrating its efficacy in tackling complex regression problems where uncertainty is a key quantity of interest.
	\end{abstract}

	\section{Introduction}
	Recent advances in Machine learning (ML) and, in particular, in Deep Learning (DL) have enabled groundbreaking advancements in modeling intricate, multi-patterned phenomena (see, e.g.,~\citet{2019zhengSelfSupervisedMixtureofExpertsUncertainty, 2021hazimehDSelectkDifferentiableSelection, 2022pontiCombiningModularSkills, 2022fedusSwitchTransformersScaling, 2023zhaoCollaborativeMixtureofExpertsModel}).
	However, these impressive achievements have often relied on increasingly complex models, hindering the tasks of \emph{quantifying uncertainty} and \emph{explaining} the predictive machinery (see~\citet{2021abdarReviewUncertaintyQuantification, 2021burkartSurveyExplainabilitySupervised}). Difficulty in performing Uncertainty Quantification (UQ) and interpreting ML models represents one of the major obstacles to their usage, especially in safety-critical applications such as (but not limited to) healthcare, finance, aeronautics, bio-informatics,
	automotive and traffic control (see, e.g.,~\citet{2022tambonHowCertifyMachine, 2021burkartSurveyExplainabilitySupervised}).

	\paragraph{Problem Statement.}

	In these scenarios, uncertainty quantification of both parameters -- the \enquote{decision-making logic} within the model -- and predictions -- the \enquote{decision} itself -- is a determinant factor in building \emph{trust} on the learned model. The latter is also often conditional on interpretability, and sheer performance may not ultimately be enough for a model to be truly of use. The presence of multiple generating mechanisms or operating regimes further limits the effectiveness of strategies such as \textit{post-hoc} explainability techniques, highlighting the importance of interpretability by design (see~\citet{2021burkartSurveyExplainabilitySupervised}).
	In light of these considerations, the Bayesian framework provides a plethora of techniques to attach uncertainty (i.e., a probability density) to estimates and predictions (see~\citet{1995draperAssessmentPropagationModel, 2009bernardoBayesianTheory, 2013zyphurBayesianEstimationInference, 2020gelmanBayesianWorkflow}), describing conditional distributions of the response variables rather than focusing only on specific functionals (see, e.g.,~\citet{2023wadeBayesianDependentMixture}).
	Nonetheless, in view of the computational effort of Bayesian Inference (as discussed in~\citet{2020luengoSurveyMonteCarlo, 2022gunapatiVariationalInferenceAlternative}), it is convenient to rely on architectures where complexity is \enquote{confined} to where strictly necessary, improving interpretability as well as scalability.
	\paragraph{Our contribution.}
	In the spirit of \textit{parsimony}, as well as \textit{expressiveness} and \emph{interpretability}, we propose to overcome these limitations with a generalization of Bayesian MoEs with finite experts, named Competitive/CollAborative Fusion of Experts (CoCoAFusE). Instead of relying on the sole ``divide et impera'' principle guiding classical MoEs, we propose a Bayesian formulation that alternates between the latter and a \textit{viribus unitis} (\enquote{joint forces}) strategy depending on the covariates, as explained in Section~\ref{sec:cocoafuse}. The key motivation for embracing CoCoAFusE is the fact that observed phenomena might not only consist of a juxtaposition of different stochastic processes but, depending on some covariates, they might also exhibit smooth transitions from one generating mechanism to another. This is indeed true in applications related to finance, physics, dynamical systems theory, and anomaly detection (see e.g.,~\citet{2024aydinhanIdentifyingPatternsFinancial, 2021callahamLearningDominantPhysical, 2003lundberghTimeVaryingSmoothTransition, 2015iamsumangHybridDBNMonitoring, 2020huAnomalyDetectionRemaining}) just to mention a few.
	We encode this feature as \textit{blends} of the sub-models, as opposed to mixtures.
	Through a suite of illustrative and real test cases, we prove that CoCoAFusE can offer significantly tighter confidence bounds, steering clear of the MoE's shortcomings in the presence of soft transitions while being as effective as the traditional MoE under abrupt pattern changes. At the same time, we compare our solution to its blending-only counterpart (the \enquote{Blend of Experts}, or BoE).

	\section{Preliminaries}\label{sec:preliminaries}

	Over the course of this section, we aim to provide a very simple introduction to some key concepts in Bayesian Statistics that will be useful in the following.
	Of course, this is in now way an exhaustive introduction to the topic, for which we refer the interested reader, e.g., to~\citet{2021vandeschootBayesianStatisticsModellinga}.

	\subsection{The Framework of Bayesian Inference}\label{ssec:bayesian-workflow}

	The Bayesian Framework has garnered widespread attention in recent years,
	as many inference techniques were previously limited by the high computational costs.
	Bayesian tools are needed now more than ever before, in light of increasing popularity of complex applications and large datasets, begging questions regarding the appropriate treatment of uncertainties associated to data generation and model specification.
	While the Bayesian approach is not exempt from limitations and critiques, it unarguably offers an extremely elegant way to tackle many of the aforementioned problems by means of the Bayes theorem.

	\begin{theorems}[Bayes' Theorem]\label{th:bayes}
		The Bayes' rule ties the conditional probabilities of two events $A$ and $B$ via the relationship
		\begin{equation}
			P(A | B) = P(B | A) P(A) / P(B);
		\end{equation}
		the previous can be extended to the case where $A$ and $B$ represent statements on data and model parameters, i.e.,
		\begin{equation}\label{eq:bayes-theorem}
			P(\bm{\theta} | \bm{y}) = P(\bm{y} | \bm{\theta}) P(\bm{\theta}) / P(\bm{y}),
		\end{equation}
		where $\bm{y}$ represents observed data and $\bm{\theta}$ a set of model parameters.
	\end{theorems}

\begin{figure}[b!]
		\centering
		\includegraphics[width=\textwidth, trim={0 0 0 10cm}, clip]{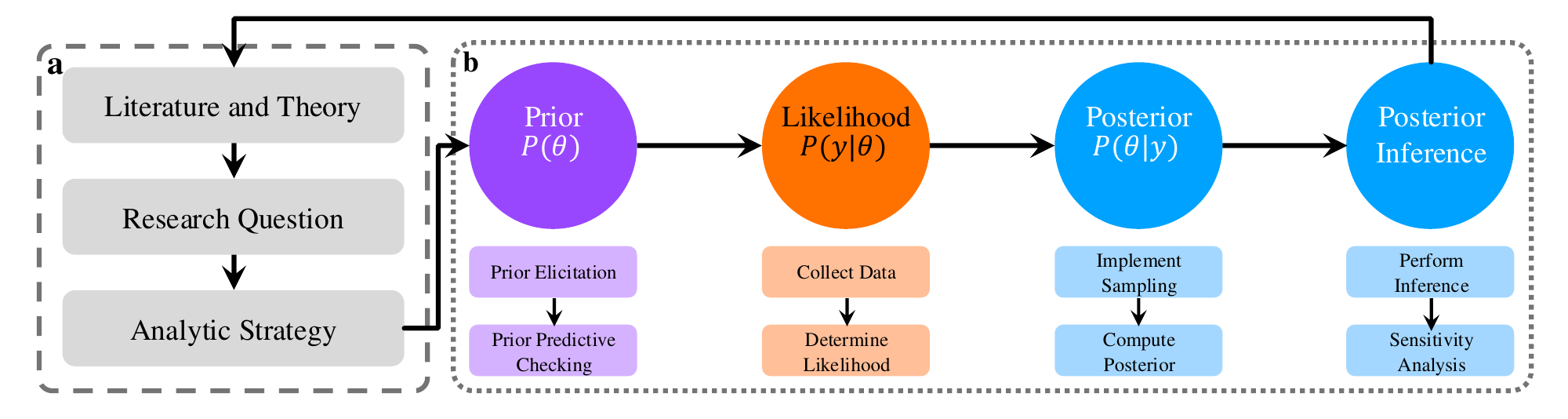}
		\caption{The Bayesian research cycle (adapted from~\citet{2021vandeschootBayesianStatisticsModellinga}). \textbf{a.} The standard research cycle. \textbf{b.} The Bayesian-specific workflow, involving prior elicitation, choice of likelihood, obtaining the posterior, and performing inference.}\label{fig:bayesian-research-cycle}
	\end{figure}

	A unique feature to Bayesian statistics is that both parameters and observed data are assigned their own probability distributions.
	The formers' is termed the \textit{prior distribution}, i.e., a probability distribution representing our beliefs on the possible values for the parameters prior to conducting any experiment.
	The latter's, instead, is generally a \textit{conditional} probability distribution for the observed quantities \textit{given} the model parameters, which is called the \textit{likelihood function}.
	\begin{remarks}[Prior, Likelihood, and Posterior Distributions]
		The expressions in equation~\eqref{eq:bayes-theorem} are the fundamental quantities in Bayesian statistics:
		$P(\bm{y} | \bm{\theta})$ is the conditional probability of the data of the data with respect to a choice $\bm{\theta}$ for the model's parameters, which is called the likelihood function of the model;
		$P(\bm{\theta})$ represents the probability of a specific set of model parameters, which is known as the prior distribution;
		$P(\bm{\theta} | \bm{y})$ is a conditional probability for the set of parameters on the observed data $\bm{y}$, which is called the posterior probability of $\bm{\theta}$.
		Notice that~\eqref{eq:bayes-theorem} involves $P(\bm{y})$ only as a normalization factor (i.e., not as a function of $\bm{\theta}$), which is why the term is often neglected replacing~\eqref{eq:bayes-theorem} with
		\begin{equation}
			P(\bm{\theta} | \bm{y}) \propto P(\bm{y} | \bm{\theta}) P(\bm{\theta}).
		\end{equation}
	\end{remarks}
	Following the Bayesian interpretation, the prior contains all knowledge that can be conveyed on the model \textit{before} observing the data.
	The likelihood, instead, postulates how a model, characterized by a specific set of parameters, affects the data generation process.
	The cornerstone of Bayesian analysis lies in applying the Bayes' Theorem~\ref{th:bayes} to combine the information contained in the priors and the observed data into a new distribution for the parameters.
	Theorem~\ref{th:bayes} then describes how do so upon collecting a set of observations, i.e, how to \textit{update knowledge} about the possible generating mechanisms that could have given rise to the data.
	The typical Bayesian workflow, therefore, can be coarsely outlined by the steps in Figure~\ref{fig:bayesian-research-cycle} (based on~\citet{2021vandeschootBayesianStatisticsModellinga}): a set of reasonable priors are formulated based on previous knowledge on the topic, followed by the definition of a suitable likelihood function for the model's parameters;
	finally, through an appropriate methodology, the resulting posterior distribution for the parameters is computed (or approximated) as per Theorem~\ref{th:bayes}, which ultimately allows inference on new data.
	In this work we define an original model (in Section~\ref{sec:cocoafuse}), which involves the deduction of a likelihood function, based on the ideas exposed in Section~\ref{ssec:mixing-blending-fusing}.
	We will briefly touch on some possible choices for the priors, adopted in the experiments (Sections~\ref{sec:didactic-examples} and~\ref{sec:experiments}), which are however not the only possible ones.
	We adopt an off-the-shelf algorithm for the sampling from the posterior distribution (the Stan\footnote{Stan Modeling Language Users Guide and Reference Manual, \texttt{https://mc-stan.org}} package).

	\paragraph{Modeling Uncertainty.}

	Through the previously discussed framework(s), Bayesian practitioners can draw upon a vast array of knowledge and methods to develop tailored models and computational strategies.
	The main advantage lies in naturally addressing
	various forms of uncertainty affecting the data generation process.
	This is because the Bayesian paradigm rests on the association of probability distribution to quantities of interest, which we can interpret as a continuous \textit{measure of uncertainty} attached to them.
	Additionally, in a Bayesian setting, such uncertainties are \textit{updated} with each observation, which lends itsef to a very simple interpretation: our knowledge about the phenomena under investigation is transformed with each piece of collected data.
	This ``subjective'' approach to modeling (in the sense that it depends on a specific starting point) makes the Bayesian perspective extremely amenable in problems where uncertainty is a primary focus, setting it apart, e.g., from typical Machine Learning tasks that are more accuracy-oriented.

	\subsection{Density Regression}\label{ssec:density-regression}

	In the typical regression setting, we aim at approximating a relationship between a set of covariates $\bm{x}$ and a response variable $y$.
	Such relationship is usually intended as stochastic, meaning that
	knowledge of $\bm{x}$ does not fully characterize $y$, but rather a (conditional) probability distribution for its possible values.
	The standard statistical tool in this context is the linear regression model,
	which however imposes very strong constraints on the type of relationships between covariates and response that can be modeled.
	A more flexible approach involves postulating that the regression function is a linear combination of some \textit{basis functions};
	many powerful statistical tools, such as Kernel Regression and Gaussian Processes (see Remark~\ref{remark:benchmark-models}), can in fact be represented this way.

	\begin{remarks}[Noteworthy Examples]\label{remark:benchmark-models}
		As mentioned, Gaussian Processes Regression (GPR) can be represented as a linear regression on a set of nonlinear basis functions (see~\citet{2008rasmussenGaussianProcessesMachine}).
		We consider this approach as a performance baseline in light of its great flexibility as well as its popularity in Bayesian literature.
		We also include Bayesian Neural Networks (BNNs, see~\citet{2022jospinHandsOnBayesianNeural}) in comparisons with our proposed approach.
		The reason is twofold: on one hand BNNs represent an attempt to include rigorous Uncertainty Quantification methods within Deep Learning frameworks, while on the other they consist of inherently nonlinear regression techniques, unlike GPR.
	\end{remarks}

	A possible issue with such regression approaches is that they are usually only flexible in terms of the regression \textit{function} (see~\citet{2023wadeBayesianDependentMixture}).
	Instead, the term \textit{density regression} designates the statistical techniques that aim to directly characterize the response variable's conditional law, as opposed to one (or a few) of its \textit{functionals}.
	Density regression techniques can be broadly categorized into \textit{joint} and \textit{conditional} approaches, according to whether the joint distribution of covariates $\bm{x} \in \mathbb{R}^{n}$ and target $\bm{y}$ or the conditional distribution of $\bm{y} \in \mathbb{R}^{p}$ on $\bm{x}$ is of interest (for a review, see~\citet{2023wadeBayesianDependentMixture}).
	In this work we solely focus on the latter case.

	\paragraph{Conditional Dependent Mixture Models (CDMMs).}
	Among conditional density regression techniques at the analyst's disposal, Conditional Dependent Mixture Models (CDMMs) are noteworthy for balancing smoothness and flexibility (see~\citet{2023wadeBayesianDependentMixture}). By postulating a conditional mixture of different processes at each covariate, CDMMs are particularly suited to capture local behaviors. Meanwhile, the regularity of the final CDMM density is determined via the definition of appropriate maps from the covariates to the mixtures.
	When the number of processes (i.e., experts) is countable, CDMMs can be restated in terms of latent discrete allocations to the experts, with their interpretation as a stochastic selection of an appropriate sub-model, followed by a prediction from the latter.

	\subsection{Mixing, Blending, and Everything in Between}\label{ssec:mixing-blending-fusing}

		\paragraph{Mixing.}

		As mentioned towards the end of the previous Section~\ref{ssec:density-regression}, a noteworthy family of Density Regression techniques assumes that the response is, conditionally on a set of regressors, a \textit{mixture} of a (countable) set of base densities.
		More formally, given a countable family of probability densities $\mathcal{F} = \{f_i(x)\}_{i \in S}$ with $S\subset\mathbb{N}$, and a sequence of non-negative scalars $\boldsymbol{\pi} = \{\pi_i\}_{i \in S}$ such that $\sum_{i \in S}\pi_i = 1$, the ``mixture'' of densities $\mathcal{F}$ with weights $\boldsymbol{\pi}$ is simply defined as the density
		\begin{equation}\label{eq:mixture-def}
			f_{\boldsymbol{\pi}}^{\textup{mix}}(x) = \sum_{i \in S}\pi_i f_i(x),
		\end{equation}
		i.e., a convex combination of the base densities.
		From here on, we will only consider \textit{finite} mixtures, i.e., with $S$ a finite set.
		Consider thus a mixture density such as in~\eqref{eq:mixture-def} where $S = \{1\dots M\}$ for some $M\in\mathbb{N}$.
		A perhaps trivial observation is that, in this scenario, $f_{\boldsymbol{\pi}}$ corresponds to the density of the following random variable:
		\begin{equation}\label{eq:mixture-as-competitive-sum}
			 X_{\textup{mix}} = \sum_{i\in S} h_i X_i,\quad h_i = \delta_{iZ},\quad Z\sim\textup{Cat}(\boldsymbol{\pi}),\quad X_i\stackrel{\textup{ind.}}{\sim} f_i,\quad \left(X_i\right)_{i\in S}\indep Z,
		\end{equation}
		i.e., of a weighted sum of independent draws from each base density $f_i$ where a single weight is non-zero, with corresponding probability $\pi_i$.
		Of course, we could have described the same random variable in a much more compact way as $\boldsymbol{X}_Z$ (to be intended as a $Z$-indexing operation on the random vector $\boldsymbol{X} = \left(X_i\right)_{i\in S}$) under an identical set of assumptions which is, in fact, how we shall deduce the likelihood expression in Section~\ref{ssec:mixtures-of-experts}.

		However, the expression derived in~\eqref{eq:mixture-as-competitive-sum} should convey why we call mixtures ``competitive'', as any draw from the mixture density~\eqref{eq:mixture-def} is morally a draw from one of the base densities $f_i$, each ``competing'' to be extracted with its own selection probabilities $\pi_i$.
		In this light, the mixture of density is the result of a repeated ``winner-takes-all'' game of the base densities.
		While this can be a powerful tool to model overlapping patterns in the response, mixture densities are generally multi-modal even when the base densities are not.
		What we will argue over the next few paragraphs is that mixture densities are extremely useful to describe complex phenomena, but only represent the extremum in a spectrum of possible ways in which independent machanisms can coalesce.

		\paragraph{Blending.}

		As opposed to mixing, we aim at definining a different way of combining the experts' densities, which an extension to the density interpolation approach defined in~\citet{1996bursalInterpolatingProbabilityDistributions}.
		We shall assume that the base densities $f_i$ all have the same shape, and are obtained as $f_i = f(\bullet; \bm{\theta}_i)$ from some parametric family $f$.
		Moreover, we assume that each $f_i$ can be appropriately characterized by its mean $\mu_i$ and variance $\sigma_i^2$ (i.e., there exists a one-to-one mapping between $\bm{\theta}_i$ and $(\mu_i, \sigma_i^2)$).
		The desired new density, combining the characteristics of each expert as a function of $\bm{\pi}$, should achieve mean and variance equal to
		\begin{subequations}\label{eq:blend-statistics}
			\begin{align}
				 \mublend &=  \sum_{i \in S}\pi_i\mu_i,\label{eq:blend-mus}\\
				 \sigmablend^2 &=  \sum_{i \in S}\pi_i\sigma_i^2\label{eq:blend-sigmas}
			\end{align}
		\end{subequations}
		(see~\citet{1996bursalInterpolatingProbabilityDistributions}).
		The reasoning behind~\eqref{eq:blend-statistics} is that, as $\bm{\pi}$ progressively lays more mass on a single expert $i$, the mean and variance of the blend should tend to $\mu_i$ and $\sigma_i^2$ respectively.

		\begin{remarks}[Interpolation of Variances]
			Although the fact that the blend's mean~\eqref{eq:blend-mus} should be a convex combination of the experts' means is rather easy to interpret, combining the variances into~\eqref{eq:blend-sigmas} could be a more dubious choice, e.g., as opposed to doing the same with the standard deviations;
			we refer the interested reader to~\citet{1996bursalInterpolatingProbabilityDistributions} (Section 2.1), for a convincing example that~\eqref{eq:blend-sigmas} is a preferrable choice.
		\end{remarks}

		So far we have not defined the expression for the blend density itself, among infinitely many with first two moments compatible with~\eqref{eq:blend-statistics}.
		Extending the approach outlined in~\citet{1996bursalInterpolatingProbabilityDistributions}, we start by definining a set of transformed coordinates $x_1,\dots,x_M$ fulfilling
		\begin{equation}
			\frac{x - \mublend}{\sigmablend} = \frac{x_1 - \mu_1}{\sigma_1} = \dots = \frac{x_M - \mu_M}{\sigma_M}.
		\end{equation}
		Observe that the Jacobians of such transformations are $\mathrm{d}x_i/\mathrm{d}x = \sigma_i/\sigmablend > 0$.
		We thus formally define the \textit{blend} as the density
		\begin{equation}\label{eq:blend-definition}
			 f_{\bm{\pi}}^{\textup{blend}}(x) = \sum_{i=1}^M \pi_i \frac{\sigma_i}{\sigmablend} f_i(x_i) = \sum_{i=1}^M \pi_i \frac{\sigma_i}{\sigmablend} f(x_i; \mu_i, \sigma_i^2).
		\end{equation}
		The following two lemmas extend the results in~\citet{1996bursalInterpolatingProbabilityDistributions}, justifying our approach.
		Proofs are contained in Appendix~\ref{app:proofs}.
		\begin{lemmas}\label{lemma:blend-is-a-density}
			The blend~\eqref{eq:blend-definition} is a valid probability density.
		\end{lemmas}
		\begin{lemmas}\label{lemma:blend-of-gaussians}
			The blend of proper Gaussians is a proper Gaussian density with parameters~\eqref{eq:blend-statistics}.
		\end{lemmas}
		We consider the \textit{blending} $f_{\bm{\pi}}^{\textup{blend}}$ to be the ``collaborative'' counterpart to~\eqref{eq:mixture-as-competitive-sum}, in the sense that all base densities' parameters contribute to the final draw.
		More specifically, each ``expert'' contributes in a weighted fashion to the mean and variance of the blend.
		In the following, we focus on $f_i = \mathcal{N}(\mu_i, \sigma_i)$, although the methodology can in principle be extended to any set of densities $f_i(x) = f(x; \mu_i, \sigma_i^2)$.

		\paragraph{Fusing.}

\begin{figure}[tb!]
			\centering
			\includegraphics[width=0.8\textwidth, trim={0 0 0 1cm}, clip]{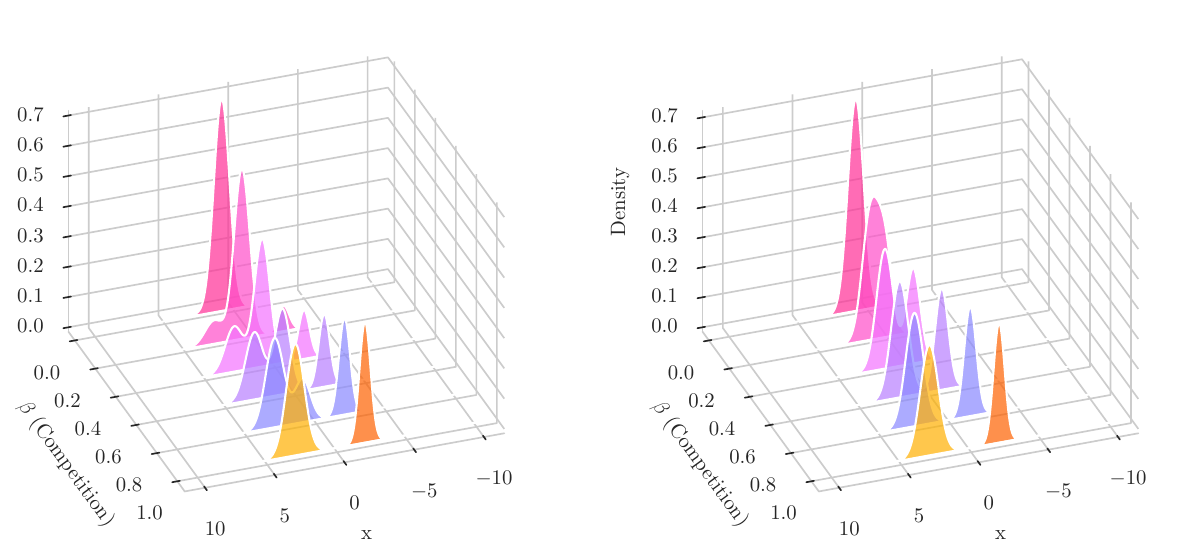}
			\caption{Two interpolation approaches between the mixture (forefront) and blend density (back). Left: the probability distribution for each $\beta$ is a na\"ive interpolation of mixture and blend, resulting in intermediate densities with an additional mode. Right: the parameters of the base densities are interpolated between the extreme cases as a function of $\beta$ and then mixed, preserving the number of modes.}\label{fig:fusion-interpolations}
		\end{figure}

		Through~\eqref{eq:mixture-as-competitive-sum} and~\eqref{eq:blend-definition} we have only introduced the \textit{extrema} of a wide range of possible generating mechanisms involving multiple (yet finitely many) individual experts $f_i$.
		We now want to describe a range of possible ``intermediate'' behaviours, which we shall indicate as \textit{fusions} for lack of better terminology (we will offer a formal definition later).
		We aim to recover them by means of a suitable \textit{interpolation} between the two extreme cases $f_{\boldsymbol{\pi}}^{\textup{mix}}$ and $f_{\boldsymbol{\pi}}^{\textup{blend}}$, in the special case of Normal densities for the individual subprocesses.
		\begin{remarks}[Na\"ive Interpolation between Densities]
			We might be drawn to directly set
			\begin{equation}\label{eq:naive-fusion}
				X_{\beta} \sim \beta f_{\textup{mix}} + (1 - \beta) f_{\textup{blend}}.
			\end{equation}
			This, however, simply results in a mixture with an additional base density $f_{\textup{blend}}$ and a set of mixture weights $(\beta\boldsymbol{\pi}, 1 - \beta)$.
		\end{remarks}
		The na\"ive interpolation can certainly be convenient in some scenarios.
		Since this is a mixture of $M + 1$ components, however, we can easily construct examples where this achieves exactly $M + 1$ modes.
		This might entail \textit{spurious} multimodalities as a result of mixing $f_{\textup{mix}}$ and $f_{\textup{blend}}$, as per the next example.
		\begin{examples}
			Consider two Gaussians
			$Y_1 \sim f_1 = \mathcal{N}(-2.0, 0.4)$,
			$Y_2 \sim f_2 = \mathcal{N}(3.0, 0.7)$,
			and mixture weights $\bm{\pi} = (0.4, 0.6)$.
			If we blend $Y_1$ and $Y_2$ according to~\eqref{eq:blend-definition}, we get $f_{\textup{blend}} = \mathcal{N}(1.0, 179/500)$.
			If we further mix $f_{\textup{mix}}$ and $f_{\textup{blend}}$ according to~\eqref{eq:naive-fusion}, we can clearly achieve a trimodal density (e.g., by setting $\beta=0.4$), as visualized in Figure~\ref{fig:fusion-interpolations} (left).
		\end{examples}

		We propose to first interpolate between the parameters of each base density and~\eqref{eq:blend-mus} and~\eqref{eq:blend-sigmas}, in a measure controlled by $\beta$, i.e., $f_i^{(\beta)} = \mathcal{N}\left(\mu_i^{(\beta)}, \sigma_i^{(\beta)}\right)$ with
		\begin{subequations}\label{eq:fusion-interpolation}
			\begin{align}
				\mu_i^{(\beta)} &=  \beta \mu_i + (1-\beta) * \mublend,\label{eq:fusion-mus}\\
				\sigma_i^{(\beta)} &=  \sqrt{\beta \sigma_i^2 + (1-\beta) * \sigmablend^2},\label{eq:fusion-sigmas}
			\end{align}
		\end{subequations}
		and only \textit{then} perform the mixture of the $f_i^{(\beta)}$ via the usual weights $\boldsymbol{\pi}$.
		\begin{lemmas}\label{lemma:interpolation-is-a-blend}
			The interpolation~\eqref{eq:fusion-interpolation} is still a blend of the base densities $\{f_i\}$ via weights $\bm{\tau}^{(i)}$.
		\end{lemmas}
		\begin{corollaries}
			As an immediate consequence to Lemma~\ref{lemma:interpolation-is-a-blend}, fusions can be easily extended to the case of general $\{f_i(x; \mu_i, \sigma_i^2)\}_{i=1}^M$ with identical shape by constructing $i$ individual blends.
		\end{corollaries}
		Formally, hence, fusions are $\bm{\pi}$-mixtures of different blends, achieved through equations~\eqref{eq:fusion-interpolation}, i.e., with sets of weights $\bm{\tau}^{(i)}$ that converge to $\bm{\pi}$ as $\beta\to0$ and to $(\delta_{ji})_{j=1}^M$ as $\beta\to1$.
		\begin{propositions}\label{prop:number-of-modes}
			The Fusion of $M$ proper Gaussians in 1D has at most $M$ modes.
		\end{propositions}
		Because of Proposition~\ref{prop:number-of-modes}, in the special case of Gaussian base densities and a scalar response $y$, the number of modes of the fusion can never be larger than $M$, resulting in no ``spurious'' multimodalities.
		The result of this approach is portrayed in Figure~\ref{fig:fusion-interpolations} (right): as $\beta$ approaches 0, all the base mixtures are translated and scaled to the same blend density.
		\begin{remarks}
			Notice that Proposition~\ref{prop:number-of-modes} does not hold in higher dimension.
			For instance, see Example 2.2 in~\citet{2020amendolaMaximumNumberModes}.
		\end{remarks}

	\subsection{Mixtures of Experts}\label{ssec:mixtures-of-experts}

	The Mixture of Experts (MoE) introduced by~\citet{1991jacobsAdaptiveMixturesLocal} is an associative competitive learning model in which several learners (\textit{experts}) are combined by a \emph{gate} to solve supervised learning problems.
	The MoE allows to achieve high flexibility while maintaining substantial local explainability, with the gate enhancing expressiveness without sacrificing the simplicity of the experts (see~\citet{2006kanaujiaLearningAmbiguitiesUsing, 2012yukselTwentyYearsMixture, 2017baldacchinoRobustNonlinearSystem}).
	The structural choices made with the MoE can be seen as a way to shift complexity from the experts onto the gate by assigning each expert to a simpler sub-problem.
	Ultimately, it
	means that \textit{local} explainability of the predictions can be preserved as a gated superimposition of \enquote{simple} decisions.

	Although the MoE was originally introduced by~\citet{1991jacobsAdaptiveMixturesLocal} in a \textit{frequentist} setting, several Bayesian formulations have translated it into an instance of CDMM where the mixture over a countable set of experts is conditional on the set of covariates (see, e.g.,~\citet{1995waterhouseBayesianMethodsMixtures, 2006kanaujiaLearningAmbiguitiesUsing, 2020xiaBayesianMixtureExperts}).
	Bayesian Mixtures of Experts are thus models where the final prediction is a discrete mixture of the individual experts' predictions, with weights provided by the \textit{gate} (as schematically shown in Figure~\ref{subfig:moe-scheme}).
	Because of the latent allocation interpretation from~\citet{2023wadeBayesianDependentMixture},
	MoEs can be seen as a \textit{divide et impera} approach to regression, with the gate partitioning the covariate space into sub-domains, whereby a single sub-model is (stochastically) tasked to generate predictions.
	Some notable extensions of this framework have dealt with
	hierarchies of MoEs~\citep{2012bishopBayesianHierarchicalMixtures}, alternative descriptions of the noise~\citep{2017baldacchinoRobustNonlinearSystem}, and non-parametric generalizations of the gate model~\citep{2022zhangSimilaritybasedBayesianMixtureofexperts}.
	Bayesian MoEs can also be distinguished based on the number of experts, which can be either finite, unspecified (see~\citet{1997richardsonBayesianAnalysisMixtures}), or even countably infinite (see~\citet{2001rasmussenInfiniteMixturesGaussian, 2005meedsAlternativeInfiniteMixture}). In this work, we only focus on \textit{finite} Mixtures of Experts, i.e., the number of experts $M$ is arbitrary but fixed apriori.
	We elect this class of models because it directly lets the designer constrain the complexity of the model, as well as ensure interpretability.
	Net of the peculiarities of the different approaches, the general scheme of a Bayesian MoE with finite experts boils down to the following core elements:
	\textit{(i)} a set of $M$ generation mechanisms $\mathcal{E}_1,\dots, \mathcal{E}_M$ for the target variable $\bm{y}$, conditionally on a set of predictors $\bm{x}$ and parameters $\bm{\theta}_1, \dots, \bm{\theta}_M$,
	\textit{(ii)} a \textit{gate} model $\mathcal{G}$, generating a (discrete) mixture over the set of experts conditionally on $\bm{x}$ and gate-specific parameters $\bm{\theta}_{\mathcal{G}}$,
	\textit{(iii)} priors on the uncertain parameters $\bm{\theta} = \left[\bm{\theta}_1^\top, \cdots, \bm{\theta}_M^\top, \bm{\theta}_{\mathcal{G}}^\top\right]^\top$.

\begin{figure}[tb!]
	\centering
	\subfigure[MoE]{\label{subfig:moe-plate}
		\adjustbox{height=0.12\textheight, trim={0 0 0 0.2cm}}{
			\begin{tikzpicture}[scale=0.5]
				\node[draw,thick,circle,fill=ibmblue!80!white,minimum size=2.5em,node distance=1.5cm] (xn) {$x_n$};
				\node[draw,thick,circle,fill=ibmblue!80!white,left of=xn,minimum size=2.5em,node distance=1.5cm] (yn) {$y_n$};
				\node[draw,thick,circle,fill=ibmpink!20!white,below of=yn,minimum size=2.5em,node distance=1.5cm] (zn) {$z_n$};
				\node[draw,thick,circle,fill=ibmpink!20!white,left of=zn,minimum size=2.5em,node distance=1.5cm] (thetai) {$\bm{\theta}_i$};
				\node[draw,thick,circle,fill=ibmpink!20!white,left of=yn,minimum size=2.5em,node distance=1.5cm] (sigmai) {$\sigma_i$};
				\node[draw,thick,circle,fill=ibmpink!20!white,below left of=zn,minimum size=2.4em,node distance=2.1213cm] (thetag) {$\bm{\theta}_\mathcal{G}$};
				\node[draw,thick,rectangle,fill=tolsand!40!white,left of=thetai,node distance=1.5cm,minimum size=7mm] (theta0i) {$\bm{\theta}^{(0)}_i$};
				\node[draw,thick,rectangle,fill=tolsand!40!white,left of=sigmai,node distance=1.5cm,minimum size=7mm] (sigma0i) {$\sigma^{(0)}_i$};
				\node[draw,thick,rectangle,fill=tolsand!40!white,left of=thetag,node distance=1.5cm,minimum size=7mm] (theta0g) {$\bm{\theta}^{(0)}_\mathcal{G}$};

				\draw[->, thick] (xn) -- (yn);
				\draw[->, thick] (xn) -- (zn);
				\draw[->, thick] (zn) -- (yn);
				\draw[->, thick] (thetai) -- (yn);
				\draw[->, thick] (sigmai) -- (yn);
				\draw[->, thick] (thetag) -- (zn);
				\draw[->, thick] (theta0i) -- (thetai);
				\draw[->, thick] (sigma0i) -- (sigmai);
				\draw[->, thick] (theta0g) -- (thetag);

				\begin{pgfonlayer}{background}
					\path (yn.west |- yn.north)+(-0.2,0.2) node (a) {};
					\path (xn.east |- zn.east)+(+0.2,-1.1) node (c) {};
					\path (zn.south -| zn.east)+(+1.9,-0.55) node (b) {\textcolor{ibmorange!70!black}{$N$}};
					\path[fill=ibmorange!15!white,thick,rounded corners, draw=ibmorange!80!black]
					(a) rectangle (c);
				\end{pgfonlayer}

				\begin{pgfonlayer}{background}
					\path (sigma0i.west |- sigmai.north)+(-0.2,0.2) node (a) {};
					\path (thetai.east |- theta0i.east)+(+0.2,-1.1) node (c) {};
					\path (thetai.south -| theta0i.east)+(-1.3,-0.55) node (b) {\textcolor{ibmpurple!70!black}{$M$}};
					\path[fill=ibmpurple!15!white,thick,rounded corners, draw=ibmpurple!80!black]
					(a) rectangle (c);
				\end{pgfonlayer}
			\end{tikzpicture}
		}
	}\hspace{0.1\textwidth}
	\subfigure[CoCoAFusE]{\label{subfig:cocoafuse-plate}
		\adjustbox{height=0.12\textheight, trim={0 0 0 0.2cm}}{
			\begin{tikzpicture}[scale=0.5]
				\node[draw,thick,circle,fill=ibmblue!80!white,minimum size=2.5em,node distance=1.5cm] (xn) {$x_n$};
				\node[draw,thick,circle,fill=ibmblue!80!white,left of=xn,minimum size=2.5em,node distance=1.5cm] (yn) {$y_n$};
				\node[draw,thick,circle,fill=ibmpink!20!white,below of=yn,minimum size=2.5em,node distance=1.5cm] (zn) {$\bm{\alpha}_n$};
				\node[draw,thick,circle,fill=ibmpink!20!white,left of=zn,minimum size=2.5em,node distance=1.5cm] (thetai) {$\bm{\theta}_i$};
				\node[draw,thick,circle,fill=ibmpink!20!white,left of=yn,minimum size=2.5em,node distance=1.5cm] (sigmai) {$\sigma_i$};
				\node[draw,thick,circle,fill=ibmpink!20!white,below left of=zn,minimum size=2.4em,node distance=2.1213cm] (thetag) {$\bm{\theta}_\mathcal{G}$};
				\node[draw,thick,rectangle,fill=tolsand!40!white,left of=thetai,node distance=1.5cm,minimum size=7mm] (theta0i) {$\bm{\theta}^{(0)}_i$};
				\node[draw,thick,rectangle,fill=tolsand!40!white,left of=sigmai,node distance=1.5cm,minimum size=7mm] (sigma0i) {$\sigma^{(0)}_i$};
				\node[draw,thick,rectangle,fill=tolsand!40!white,left of=thetag,node distance=1.5cm,minimum size=7mm] (theta0g) {$\bm{\theta}^{(0)}_\mathcal{G}$};

				\node[draw,thick,circle,fill=ibmpink!20!white,below of=xn,minimum size=2.5em,node distance=1.5cm] (wn) {$\beta_n$};
				\node[draw,thick,circle,fill=ibmpink!20!white,below of=zn,minimum size=2.4em,node distance=1.5cm] (thetab) {$\bm{\theta}_\mathcal{B}$};
				\node[draw,thick,rectangle,fill=tolsand!40!white,right of=thetab,node distance=1.5cm,minimum size=7mm] (theta0b) {$\bm{\theta}^{(0)}_\mathcal{B}$};

				\draw[->, thick] (xn) -- (yn);
				\draw[->, thick] (zn) -- (yn);
				\draw[->, thick] (xn) -- (zn);
				\draw[->, thick] (xn) -- (wn);
				\draw[->, thick] (wn) -- (yn);
				\draw[->, thick] (thetai) -- (yn);
				\draw[->, thick] (sigmai) -- (yn);
				\draw[->, thick] (thetab) -- (wn);
				\draw[->, thick] (thetag) -- (zn);
				\draw[->, thick] (theta0i) -- (thetai);
				\draw[->, thick] (sigma0i) -- (sigmai);
				\draw[->, thick] (theta0b) -- (thetab);
				\draw[->, thick] (theta0g) -- (thetag);

				\begin{pgfonlayer}{background}
					\path (yn.west |- yn.north)+(-0.2,0.2) node (a) {};
					\path (xn.east |- zn.east)+(+0.2,-1.1) node (c) {};
					\path (zn.south -| zn.east)+(+1.9,-0.55) node (b) {\textcolor{ibmorange!70!black}{$N$}};
					\path[fill=ibmorange!15!white,thick,rounded corners, draw=ibmorange!80!black]
					(a) rectangle (c);
				\end{pgfonlayer}

				\begin{pgfonlayer}{background}
					\path (sigma0i.west |- sigmai.north)+(-0.2,0.2) node (a) {};
					\path (thetai.east |- theta0i.east)+(+0.2,-1.1) node (c) {};
					\path (thetai.south -| theta0i.east)+(-1.3,-0.55) node (b) {\textcolor{ibmpurple!70!black}{$M$}};
					\path[fill=ibmpurple!15!white,thick,rounded corners, draw=ibmpurple!80!black]
					(a) rectangle (c);
				\end{pgfonlayer}
			\end{tikzpicture}
		}
	}
	\caption{Plate Diagrams for the MoE and CoCoAFusE. Blue shaded circles represent observed data, light pink shaded circles indicate uncertain parameters, sand rectangles are prior hyperparameters, and plates represent sub-graphs that are repeated (plate $M$ corresponds to the various experts, and plate $N$ corresponds to data points). Arrows represent conditional dependence.}
\end{figure}
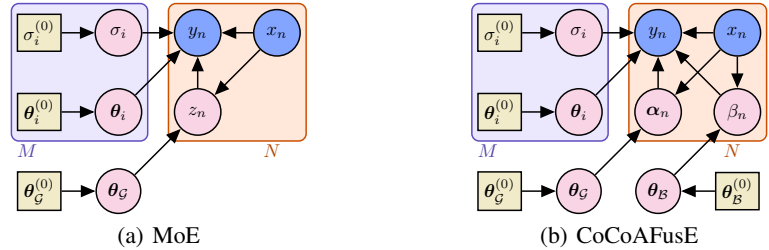

	By introducing the categorical variable $z\in\{1,\dots,M\}$, denoting the (latent) allocation to expert $i\in\{1, \dots,M\}$, and the associated probability $p(z = i | \bm{x};\: \bm{\theta}_{\mathcal{G}})$ conditional on $\bm{x}\in\mathbb{R}^n$ and $\bm{\theta}_{\mathcal{G}}$, the target's likelihood conditional on $\bm{x}$ and $\bm{\theta}$ can thus be broken down as:
	\begin{equation}\label{eq:moe-likelihood}
		 p(\bm{y} | \bm{x}; \bm{\theta}) = \sum_{i=1}^M p(\bm{y} | \bm{x}; z = i, \bm{\theta}) p(z = i | \bm{x}; \bm{\theta}) = \sum_{i=1}^M p(\bm{y} | \bm{x}; \bm{\theta}_i) p(z = i |  \bm{x};\: \bm{\theta}_{\mathcal{G}}).
	\end{equation}
	\begin{remarks}[Mixture of Conditional Gaussian Experts]\label{remark:gaussian-conditional-experts}
		Although MoEs can be
		extremely general as to what are the experts' conditional densities $p(\bm{y} | \bm{x}; z = i, \bm{\theta}) = p\left(\bm{y} | \bm{x}; \bm{\theta}_i\right)$, we will restrict our focus to conditionally Gaussian densities on the covariates $\bm{x}\in\mathbb{R}^n$ of the form:
		\begin{equation}\label{eq:gaussian-expert-moe}
			p\left(\bm{y} | \bm{x}; \bm{\theta}_i\right) = \mathcal{N} \left(\theta_{i, 0} + \bm{x}^\top\bm{\theta}_{i, 1:n}, \theta_{i, n+1} \right).
		\end{equation}
		In this parameterization, each expert is fully specified via a parameter vector $\bm{\theta}_i\in\mathbb{R}^{n+2}$, of which the first $n+1$ components specify intercept and (linear) feature coefficients, whereas the last encodes the residual standard deviation of generating mechanism $i$.
	\end{remarks}
	\begin{remarks}[Parameterization of the Gate]\label{remark:gate-parameterization}
		When the regressor $\bm{x}$ is multivariate, a possible choice for the gate probabilities $p(\bm{z} | \bm{x};\: \bm{\theta}_{\mathcal{G}})$ is to consider a categorical distribution
		\begin{equation}\label{eq:cat_distr}
			z | \bm{x};\: \bm{\theta}_{\mathcal{G}}\sim \textup{Cat}(h(\bm{x};\bm{\theta}_{\mathcal{G}})),
		\end{equation}
		where $h:\mathbb{R}^n \to \Delta^M$ is a user-defined function shaping the distribution based on the covariates. In all our tests, we impose
		\begin{equation}\label{eq:gate}
			h(\bm{x};\bm{\theta}_{\mathcal{G}})=
			\textup{softmax}(\bm{\theta}_{\mathcal{G}}^\top\Phi(\bm{x})),
		\end{equation}
		with $\Phi:\mathbb{R}^n \to \{1\}\times\mathbb{R}^{q-1}$ an embedding function and $\bm{\theta}_{\mathcal{G}}$ a matrix in $\mathbb{R}^{q\times M}$.
		This yields a parameterization for the gates upon which priors can be elicited.
	\end{remarks}
	A plate diagram of the model represented in equation~\eqref{eq:moe-likelihood} is shown in Figure~\ref{subfig:moe-plate}.

	\paragraph{Limitations of Bayesian Mixtures of Experts.}
	Despite their many advantages in a ``multi-patterned'' setting, MoEs ultimately describe the target as a (conditional) mixture of several -- yet, finitely many -- distinct densities corresponding to the individual experts.
	Aside from uncertainties associated with the expert's choice and the response functions, observations still fall into the description of one (and one only) among the experts.
	This feature can severely affect the quality of the model and the corresponding predictive uncertainty when smooth transitions are possible between the regimes (as, e.g., for hybrid dynamical systems, see~\citet{2020casauHybridControlRobust, 2023moradvandiModelsMethodsHybrid}).
	Indeed, in this case, a finite number of local descriptions must be used to approximate the continuous family of conditional target densities.
	Imposing finitely many experts can then lead to redundancies and extraneous multimodalities that are difficult to interpret, in the worst case resulting in poor estimates and variance inflation, as we intend to show in Section~\ref{sec:didactic-examples} (see Figure~\ref{fig:transition-posterior-predictive}).
	A possible solution to these shortcomings lies in considering an unbounded and possibly infinite number of experts (\citet{2001rasmussenInfiniteMixturesGaussian, 2005meedsAlternativeInfiniteMixture}) at the price of losing \emph{parsimony} and part of the \emph{interpretability} of the model.

	\section{Fusing Experts: the CoCoAFusE framework}\label{sec:cocoafuse}

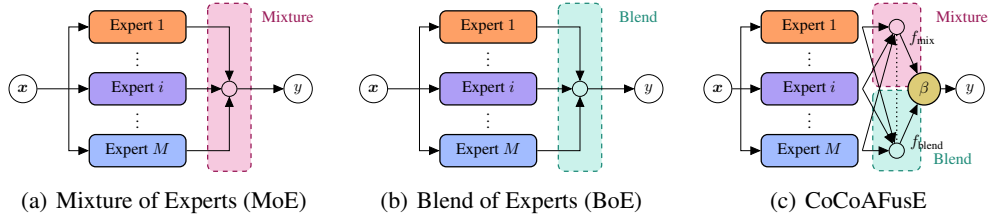
\begin{figure}[tb!]
		\centering
		\subfigure[Mixture of Experts (MoE)]{\label{subfig:moe-scheme}
			\adjustbox{height=0.1\textheight, trim={0 0.6cm 0 0.2cm}}{
				\begin{tikzpicture}[scale=0.5]
					\node[coordinate] (reference) {};
					\node[draw,circle,node distance=.75cm] (sum1) {$\bm{x}$};
					\node[coordinate,right of=sum1,node distance=1cm] (aid4) {};
					\node[draw,rectangle,thick,rounded corners,minimum height=2em,minimum width=6em,right of=sum1,node distance=2.5cm,fill=ibmpurple!60!white] (Ci) {Expert $i$};

					\node[draw,circle,right of=Ci,node distance=2cm] (sum2) {};
					\node[rectangle,above of=Ci,node distance=.75cm] (aid6) {$\vdots$};
					\node[rectangle,below of=Ci,node distance=.6cm] (aid7) {$\vdots$};
					\node[draw,rectangle,above of=aid6,node distance=.6cm,thick,rounded corners,minimum height=2em,minimum width=6em,fill=ibmorange!60!white] (C1) {Expert $1$};

					\node[draw,rectangle,thick,rounded corners,minimum height=2em,minimum width=6em,below of=aid7,node distance=.75cm,fill=ibmblue!60!white] (CN) {Expert $M$};

					\node[draw,circle,right of=sum2,node distance=1.5cm] (plant) {$y$};
					\node[coordinate,right of=plant,node distance=.75cm] (aid1) {};
					\node[coordinate,below of=CN,node distance=1cm] (aid2) {};
					\node[rectangle,below of=aid2,node distance=.1cm] (aid3) {};
					\draw[-] (sum1) -- (aid4);
					\draw[->] (aid4) -- (Ci);
					\draw[->] (aid4) |- (C1);
					\draw[->] (aid4) |- (CN);
					\draw[->] (C1) -| (sum2);
					\draw[->] (Ci) -- (sum2);
					\draw[->] (CN) -| (sum2);
					\draw[->] (sum2) -- (plant);
					\begin{pgfonlayer}{background}
						\path (sum2.west |- C1.north)+(-0.6,0.3) node (a) {};
						\path (sum2.east |- CN.east)+(+0.6,-0.9) node (c) {};
						\path (C1.south -| sum2.east)+(2.2,1.2) node (b) {\textcolor{ibmpink!70!black}{Mixture}};
						\path[fill=ibmpink!25!white,rounded corners,dashed, thick, draw=ibmpink!70!black]
						(a) rectangle (c);
					\end{pgfonlayer}
				\end{tikzpicture}
			}
		}\hspace{0.01\textwidth}
		\subfigure[Blend of Experts (BoE)]{\label{subfig:boe-scheme}
			\adjustbox{height=0.1\textheight, trim={0 0.6cm 0 0.2cm}}{
				\begin{tikzpicture}[scale=0.5]
					\node[coordinate] (reference) {};
					\node[draw,circle,node distance=.75cm] (sum1) {$\bm{x}$};
					\node[coordinate,right of=sum1,node distance=1cm] (aid4) {};
					\node[draw,rectangle,thick,rounded corners,minimum height=2em,minimum width=6em,right of=sum1,node distance=2.5cm,fill=ibmpurple!60!white] (Ci) {Expert $i$};

					\node[draw,circle,right of=Ci,node distance=2cm] (sum2) {};
					\node[rectangle,above of=Ci,node distance=.75cm] (aid6) {$\vdots$};
					\node[rectangle,below of=Ci,node distance=.6cm] (aid7) {$\vdots$};
					\node[draw,rectangle,above of=aid6,node distance=.6cm,thick,rounded corners,minimum height=2em,minimum width=6em,fill=ibmorange!60!white] (C1) {Expert $1$};

					\node[draw,rectangle,thick,rounded corners,minimum height=2em,minimum width=6em,below of=aid7,node distance=.75cm,fill=ibmblue!60!white] (CN) {Expert $M$};

					\node[draw,circle,right of=sum2,node distance=1.5cm] (plant) {$y$};
					\node[coordinate,right of=plant,node distance=.75cm] (aid1) {};
					\node[coordinate,below of=CN,node distance=1cm] (aid2) {};
					\node[rectangle,below of=aid2,node distance=.1cm] (aid3) {};
					\draw[-] (sum1) -- (aid4);
					\draw[->] (aid4) -- (Ci);
					\draw[->] (aid4) |- (C1);
					\draw[->] (aid4) |- (CN);
					\draw[->] (C1) -| (sum2);
					\draw[->] (Ci) -- (sum2);
					\draw[->] (CN) -| (sum2);
					\draw[->] (sum2) -- (plant);
					\begin{pgfonlayer}{background}
						\path (sum2.west |- C1.north)+(-0.6,0.3) node (a) {};
						\path (sum2.east |- CN.east)+(+0.6,-0.9) node (c) {};
						\path (C1.south -| sum2.east)+(2.2,1.2) node (b) {\textcolor{ibmaqua!70!black}{Blend}};
						\path[fill=ibmaqua!25!white,rounded corners,dashed, thick, draw=ibmaqua!70!black]
						(a) rectangle (c);
					\end{pgfonlayer}
				\end{tikzpicture}
			}
		}\hspace{0.01\textwidth}
		\subfigure[CoCoAFusE]{\label{subfig:cocoafuse-scheme}
			\adjustbox{height=0.1\textheight, trim={0 0.6cm 0 0.2cm}}{
				\begin{tikzpicture}[scale=0.5]
					\node[coordinate] (reference) {};
					\node[draw,circle,node distance=.75cm] (sum1) {$\bm{x}$};
					\node[coordinate,right of=sum1,node distance=0.5cm] (aid4) {};
					\node[draw,rectangle,thick,rounded corners,minimum height=2em,minimum width=6em,right of=sum1,node distance=2cm,fill=ibmpurple!60!white] (Ci) {Expert $i$};
					\node[coordinate,right of=C1,node distance=0.65cm] (epoint1) {};

					\node[coordinate,right of=Ci,node distance=1.155cm] (epointi) {};

					\node[coordinate,right of=CN,node distance=0.65cm] (epointN) {};

					\node[draw,circle,right of=C1,node distance=1.4cm] (sum2) {};
					\node[draw,circle,right of=CN,node distance=1.4cm] (sum3) {};
					\node[rectangle,above of=Ci,node distance=.75cm] (aid6) {$\vdots$};
					\node[rectangle,below of=Ci,node distance=.6cm] (aid7) {$\vdots$};
					\node[draw,rectangle,above of=aid6,node distance=.6cm,thick,rounded corners,minimum height=2em,minimum width=6em,fill=ibmorange!60!white] (C1) {Expert $1$};

					\node[draw,rectangle,thick,rounded corners,minimum height=2em,minimum width=6em,below of=aid7,node distance=.75cm,fill=ibmblue!60!white] (CN) {Expert $M$};

					\node[draw,circle,right of=Ci,thick,fill=tolsand,node distance=2.5cm] (plant) {$\beta$};
					\node[draw,circle,right of=plant,node distance=1cm] (out) {$y$};
					\node[coordinate,right of=plant,node distance=.75cm] (aid1) {};
					\node[coordinate,below of=CN,node distance=1cm] (aid2) {};
					\node[rectangle,below of=aid2,node distance=.1cm] (aid3) {};
					\draw[-] (sum1) -- (aid4);
					\draw[->] (aid4) -- (Ci);
					\draw[->] (aid4) |- (C1);
					\draw[->] (aid4) |- (CN);
					\draw[->] (epoint1) -- (sum2);
					\draw[->] (epointi) -- (sum2);
					\draw[->] (epointN) -- (sum2);
					\draw[->] (epoint1) -- (sum3);
					\draw[->] (epointi) -- (sum3);
					\draw[->] (epointN) -- (sum3);
					\draw[->] (plant) -- (out);
					\draw[dotted, thick] (sum2) -- (sum3);
					\draw[->] (sum2) -- node[yshift=.35cm, xshift=.3cm]{$f_{\textup{mix}}$}(plant);
					\draw[->] (sum3) -- node[yshift=-.4cm, xshift=.4cm]{$f_{\textup{blend}}$}(plant);
					\begin{pgfonlayer}{background}
						\path (sum2.west |- C1.north)+(-0.7,0.3) node (a) {};
						\path (sum2.east |- Ci.east)+(+0.7,0.1) node (c) {};
						\path (C1.south -| sum2.east)+(2.5,1.2) node (b) {\textcolor{ibmpink!70!black}{Mixture}};
						\path[fill=ibmpink!25!white,thick,dashed,rounded corners, draw=ibmpink!70!black]
						(a) rectangle (c);
					\end{pgfonlayer}
					\begin{pgfonlayer}{background}
						\path (sum3.west |- Ci.north)+(-0.7,-0.8) node (a) {};
						\path (sum3.east |- CN.east)+(+0.7,-0.875) node (c) {};
						\path (CN.south -| sum3.east)+(2.1,0.4) node (b) {\textcolor{ibmaqua!70!black}{Blend}};
						\path[fill=ibmaqua!25!white,thick,dashed,rounded corners, draw=ibmaqua!70!black]
						(a) rectangle (c);
					\end{pgfonlayer}
				\end{tikzpicture}
			}
		}
		\caption{Schematic overview of the MoE, BoE, and the CoCoAFusE. Blocks of the same color are identical in the two models, with dotted lines representing the sharing of parameters. In the MoE, multiple distributions are mixed via a gate (pink dashed rectangle). In the BoE, the experts are first blended, and then predictions are generated (aqua dashed rectangle). In CoCoAFusE, we ``fuse'' experts -- mix different \textit{blends} of their densities --, as a function of the behavior gate $\beta$'s output.}
	\end{figure}

	Because of the previous arguments, the \enquote{winner takes all} perspective underpinning the gate model of MoEs is not universally applicable. Our strategy in tackling this issue is to provide an extension of the MoE, called Competitive/CollAborative Fusion of Experts (CoCoAFusE), where the number of sub-models is still bounded, but some additional expressiveness can be recovered via the \textit{fusion} mechanism touched in Section~\ref{ssec:mixing-blending-fusing}.
	The key element inspiring the definition of the CoCoAFusE is the intention to better capture intermediate behaviors between experts, especially when these have meaning to the analyst. This is true, for example, in previously-mentioned hybrid systems where a key task is to distinguish between discontinuous and smooth behaviors, as discussed in~\citet{2023moradvandiModelsMethodsHybrid}.
	To do so, we introduce an additional ``behavior'' gate structure (depicted in Figure~\ref{subfig:cocoafuse-scheme}).
	The behaviour gate outputs a scalar  $\beta \in (0, 1)$ which controls the fusion between blending ($\beta\to0$) and mixing ($\beta\to1$) as per~\eqref{eq:fusion-interpolation}.
	The behaviour gate is parameterized with $\bm{\theta}_{\mathcal{\mathcal{B}}}\in\mathbb{R}^q$ via:
	\begin{equation}
		\textup{logit}(\beta) = \ln\left({\beta}\big/{(1-\beta)}\right) = \bm{\theta}_{\mathcal{B}}^\top \Phi(\bm{x}).
	\end{equation}
	Hence, conditionally on the behaviour $\beta$ and in a similar way to the parameterization~\eqref{eq:gaussian-expert-moe}, each expert now has density
	\begin{subequations}\label{eq:cocoa-fused-experts}
		\begin{gather}
			 p\left(\bm{y} | \bm{x}; \bm{\theta}, z=i, \beta\right) = p\left(\bm{y} | \bm{x}; \bm{\theta}_i^{(\beta)}\right) = \mathcal{N} \left(\theta_{i, 0}^{(\beta)} + \bm{x}^\top\bm{\theta}_{i, 1:n}^{(\beta)}, \theta_{i, n+1}^{(\beta)} \right),\\
			\bm{\theta}_{i, 0:n}^{(\beta)} = \beta \bm{\theta}_{i, 0:n} + (1 - \beta) \bm{\theta}^{\textup{blend}}_{0:n},\quad
			\theta_{i, n+1}^{(\beta)} = \sqrt{\beta \theta_{i, n+1} + (1 - \beta) \theta^{\textup{blend}}_{n+1}},
		\end{gather}
	\end{subequations}
	with $\bm{\theta}^{\textup{blend}}$ derived from the formulae~\eqref{eq:blend-statistics} (in particular,~\eqref{eq:blend-mus} for the first $n + 1$ vector components, corresponding to multiplicative coefficients, and~\eqref{eq:blend-sigmas} for the last component, corresponding to the scale parameter).
	The new set of densities~\eqref{eq:cocoa-fused-experts} is then mixed according to the gating probabilities $\alpha_i(\bm{x}) = p(z = i | \bm{x}; \bm{\theta})$.

	\begin{remarks}[Interpretation of CoCoAFusE]
		Because of the linearity of the conditional mean of the experts in the covariates $\bm{x}$, the expressions~\eqref{eq:cocoa-fused-experts} amount to performing a blend of the experts' base conditional distributions $p\left(\bm{y} | \bm{x}; \bm{\theta}_i\right)$ from~\eqref{eq:gaussian-expert-moe} as described in~\eqref{eq:blend-definition}.
		Thus, conditionally on a set of gating probabilities $\bm{\alpha}$ and behaviour $\beta$, CoCoAFusE yields a density for the response $y$ that is a fusion of the base densities~\eqref{eq:gaussian-expert-moe}.
	\end{remarks}

Overall, the generating mechanism is still a CDMM, as -- conditionally on a set of covariates $\bm{x}$ -- the density of $y$ is once again a finite mixture of exactly $M$ (Gaussian) densities.
Because of this, we can simply write the likelihood of this new model by slightly modifying~\eqref{eq:moe-likelihood}:
\begin{equation}\label{eq:cocoafuse-likelihood}
	 p(\bm{y} | \bm{x}; \bm{\theta}) = \sum_{i=1}^M p(\bm{y} | \bm{x}; z = i, \bm{\theta}) p(z = i | \bm{x}; \bm{\theta}) = \sum_{i=1}^M p\left(\bm{y} | \bm{x}; \bm{\theta}_i^{(\beta)}\right) p(z = i |  \bm{x};\: \bm{\theta}_{\mathcal{G}}).
\end{equation}
with $\bm{\theta}_i^{(\beta)}\in \mathbb{R}^{n+2}$ following the expressions in~\eqref{eq:cocoa-fused-experts}.
From a practical perspective this allows for parsimony in the number of processes while offering additional flexibility in the way that said processes are combined in the response, as we intend to show with the examples in Section~\ref{sec:experiments}.
Meanwhile, the approach does not induce extraneous multimodalities compared to the ones already present in mixtures of $M$ components thanks to the fusion mechanism: in light of Proposition~\ref{prop:number-of-modes}.

\begin{remarks}[A Purely Collaborative Model: the Blend of Experts]\label{remark:boe}
	We define the ``Blend of Experts'' (BoE) as the model with likelihood $p(\bm{y}|\bm{x};\bm{\theta})=p(\bm{y}|\bm{x};\bm{\theta}_{\textup{blend}})$, i.e. the \textit{restriction} of CoCoAFusE to the \textit{collaborative} behavior.
	This restricted model will be instrumental in Section~\ref{sec:didactic-examples} and later in Section~\ref{sec:experiments} to discuss the limitations of considering only one of the two ``extreme'' types of behaviour.
\end{remarks}

	\section{Motivating Examples}\label{sec:didactic-examples}
	We start our analysis with two simple numerical examples, crafted to highlight key differences between the extreme cases of the CoCoAFusE (i.e., pure competition, as enforced via the classical MoE, as well as pure collaboration, via the BoE defined in Remark~\ref{remark:boe}), as opposed to the full span of the model.
	The examples are constructed to meet the hypoteses of alternatively one or the other extremum, but not both.
	In the first case (the \enquote{Switch}), we aim to assess to what extent can CoCoAFusE accurately reproduce the same output as the MoE's when its underlying assumptions are met (i.e., a stochastic ``switch'' between different processes). In the second case, we consider a scenario characterized by a \textit{smooth} transition between one generating mechanism and the next, thus violating the structural assumptions of the MoE but respecting those of the BoE.
	The two examples are constructed to share identical \textit{conditional expectation} of the response $y$ given $x$, sharing the same number $M=2$ of underlying generating processes for the scalar response variable $y$ conditionally on a scalar covariate $x$.
	In one case, the generating processes are actually \textit{mixed} by covariate-dependent proportions, while they are \textit{blended} in the other.
	Because of this, both the MoE and CoCoAFusE are potentially able to recover the ``right'' conditional expectation in both scenarios. Nonetheless, when the focus lies in \textit{density estimation},
	these examples aim at highlighting what can be lost when superimposing different processes (as in the MoE approach) instead of taking into account the full range of possible interplays between them.
	We present some qualitative results in the following, whereas quantitative assessments can be found in Appendix~\ref{app:numerical-examples}.

	\subsection{Switch}
\begin{figure}[t!]
		\centering
		\subfigure[MoE]{\includegraphics[width=0.328\linewidth, trim={0.25cm 0 0.45cm 0.25cm}, clip]{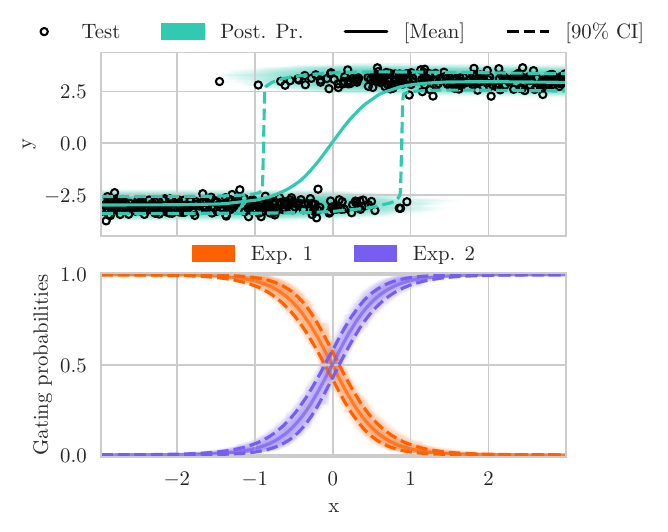}}
		\subfigure[BoE]{\includegraphics[width=0.328\linewidth, trim={0.25cm 0 0.45cm 0.25cm}, clip]{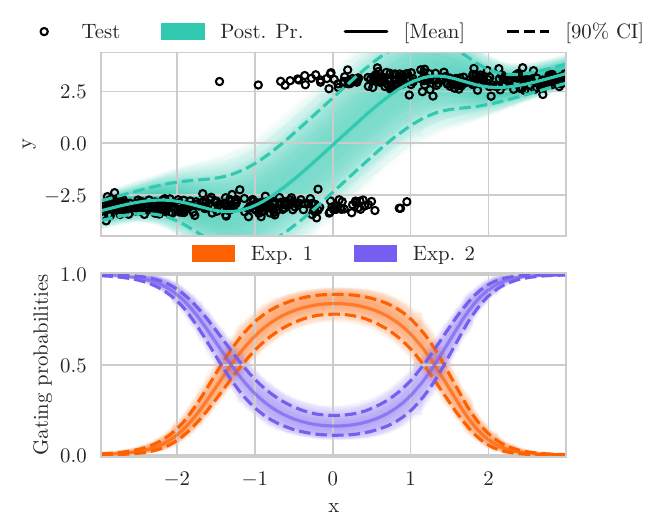}}
		\subfigure[CoCoAFusE]{\includegraphics[width=0.328\linewidth, trim={0.25cm 0 0.45cm 0.25cm}, clip]{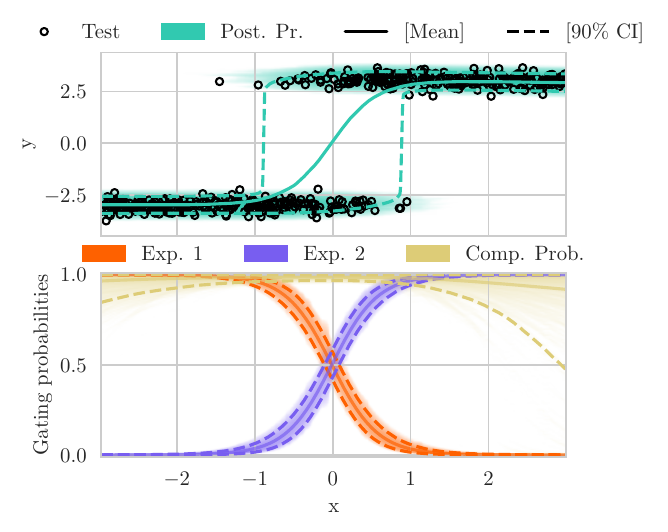}}

		\caption{Predictions for MoE, BoE, and CoCoAFusE on the Switch Example.}
		\label{fig:switch-posterior-predictive}
	\end{figure}

	We superimpose two Gaussian generating processes with equal standard deviation $\sigma = 1/4$ and means $\mu_1 = 3$ and $\mu_2 = -3$. We draw from one or the other according to a covariate-depending probability
	\begin{equation}
		p(x) = \textup{logit}^{-1} (\tau x) = 1 / 1+e^{-\tau x}.
	\end{equation}
	\begin{propositions}\label{prop:switch-conditional-mean}
		The conditional expectation of $Y$ given $X = x$ is
		\begin{equation}\label{eq:switch-conditional-mean}
			\mathbb{E}[Y|X = x] = p(x)\mu_1 + (1 - p(x))\mu_2.
		\end{equation}
	\end{propositions}
	In this scenario, we achieve near-identical results when comparing the MoE and CoCoAFuse, as confirmed by the indistinguishable posterior predictive densities shown in Figure~\ref{fig:switch-posterior-predictive}.
	Instead, the blend-only behavior enforced by the BoE results in a very poor fitting of the conditional density of the target given the input feature.
	In this example, CoCoAFusE is effective in recovering pure competition around the switch, locally behaving as the MoE, while the BoE cannot recover multimodalities that exist in the dataset (i.e., peaks in the density at \textpm3 when $x$ is near zero).

	\subsection{Smooth Transition}
\begin{figure}[tb!]
		\centering
		\subfigure[MoE]{\includegraphics[width=0.328\linewidth, trim={0.25cm 0 0.45cm 0.25cm}, clip]{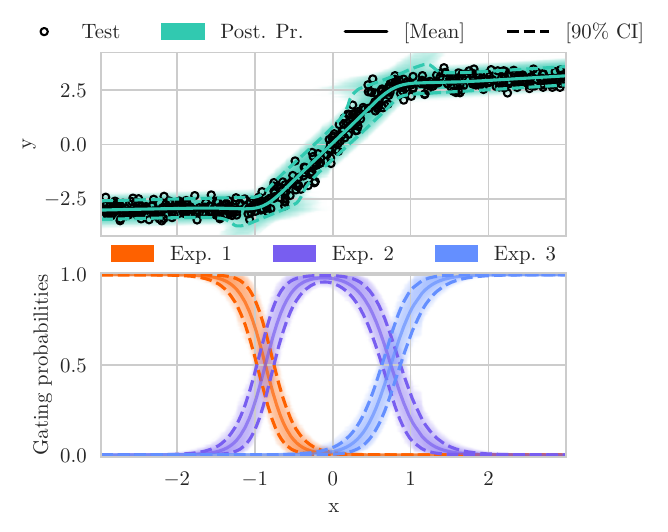}}
		\subfigure[BoE]{\includegraphics[width=0.328\linewidth, trim={0.25cm 0 0.45cm 0.25cm}, clip]{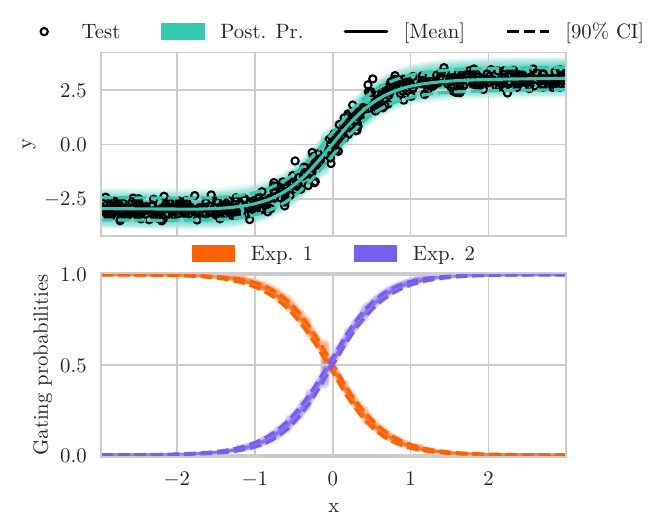}}
		\subfigure[CoCoAFusE]{\includegraphics[width=0.328\linewidth, trim={0.25cm 0 0.45cm 0.25cm}, clip]{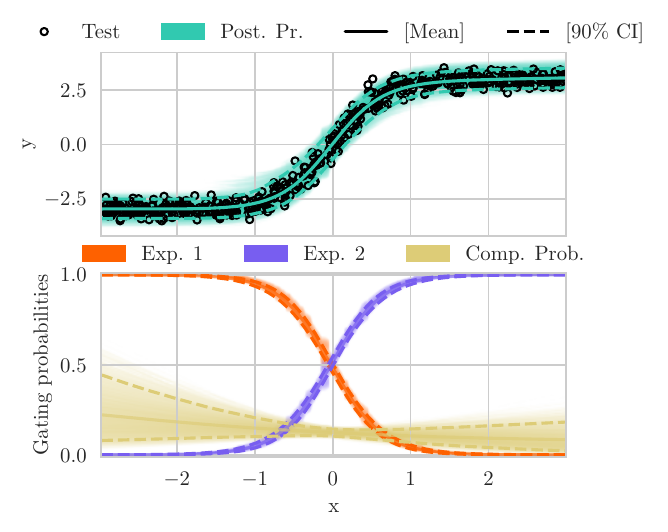}}
		\caption{Predictions for MoE, BoE, and CoCoAFusE on the Transition Example.}
		\label{fig:transition-posterior-predictive}
	\end{figure}

	We introduce two changes compared to the Switch example.
	First, we define an auxiliary variable $\alpha\in(0, 1)$ and define the conditional law hierarchically as
	\begin{subequations}\label{eq:transition-hierarchical}
		\begin{gather}
			\alpha | x \sim \textup{Beta}(k p(x), k (1 - p(x))),\ k > 0\\
			Y|\alpha, x \sim \mathcal{N}(\mu_1\alpha + \mu_2(1 - \alpha), \sigma)
		\end{gather}
	\end{subequations}
	with $\alpha$ conditionally on $x$ following a Beta distribution, parameterized via the mean $p(x)$ and total count parameter $k > 0$.
	\begin{propositions}\label{prop:transition-conditional-mean}
		The conditional mean of $Y$ given $X = x$ is exacly that of~\eqref{eq:switch-conditional-mean} for all $k > 0$.
	\end{propositions}
	Therefore, the process described in~\eqref{eq:transition-hierarchical} and the Switch example have identical conditional expectation $\mathbb{E}[Y | X = x]$ for every $k > 0$.
	If the parameter $k$ is smaller than $1$, the (conditional) distribution of $\alpha$ becomes bimodal, with the mass progressively tighter around $0$ and $1$ as $k$ decreases. Hence, in the limit for $k\to0$, $\alpha$ is approximated by a Bernoulli with parameter $p(x)$, and the Switch and Transition examples become equivalent. Conversely, for $k\to\infty$, $\alpha$ converges to the singleton $p(x)$, and the residual variance becomes uniform across the domain.
	In this case, the differences between the MoE and CoCoAFusE are stark.
	Indeed, while the correct number of experts $M = 2$ is selected by CoCoAFusE architecture, the MoE leads to a less parsimonious structure with $M = 3$ experts.
	Further, CoCoAFusE and the BoE behave similarly on this example, which underscores the presence of smooth transitions that are not captured by simply mixing linear experts. This is further visible in \figurename{~\ref{fig:transition-posterior-predictive}}, showing inflated bounds in the credible intervals of the MoE, especially in correspondence with high uncertainties on expert allocation. These undesired features are not observed for CoCoAFusE (or BoE), which shows that uncertainty over the expert selection in the MoE can lead to spurious multi-modalities, while collaboration helps in mitigating such artifacts.

	\section{CoCoAFusE, in Practice}

	\subsection{How Inference is Performed}

	Following the Bayesian workflow discussed in Section~\ref{ssec:bayesian-workflow},
	once all of the densities appearing in the likelihood and a set of priors have been specified (see the ones discussed in Appendix~\ref{app:choice-of-distr}), the posterior distribution on $\bm{\theta}$ can be computed or approximated.
	The most popular approach to do so is by Markov-Chain Monte Carlo (MCMC). We use the Stan language to implement the priors and likelihood for the model, and the built-in NUTS algorithm (from~\citet{2014hoffmanNoUTurnSamplerAdaptively}) to sample from the posterior.
	In order to use Stan, we implement the numerically robust formulations of the logarithms of~\eqref{eq:moe-likelihood} and~\eqref{eq:cocoafuse-likelihood} described in Appendix~\ref{app:numerical-robustness}.
	Following the typical Bayesian approach, the uncertain components of the model (i.e., parameters) can then be marginalized at inference time, yielding the \textit{posterior predictive density}
	\begin{equation}\label{eq:posterior-predictive-dens}
		p(\bm{y} | \bm{x}) = \int p(\bm{y} | \bm{x}; \bm{\theta}) p(\bm{\theta} | \mathcal{D})d\theta,\quad\mathcal{D} = \{(\bm{x}_n, \bm{y}_n)\}_{n=1}^{N}
	\end{equation}
	where $p(\bm{\theta} | \mathcal{D})$ is the posterior distribution of the parameters given the dataset $\mathcal{D}$.
	Since equation~\eqref{eq:posterior-predictive-dens} is generally intractable analytically, we resort to Monte-Carlo estimates.

	\subsection{Label Switching Problem of Mixture Models}\label{ssec:label-switching-problem}
	Regardless of implementation,
	conditional mixture models face issues since the
	MCMCs might ``jump'' between different permutations of the experts, rendering the chain non-stationary\footnote{M. Betancourt, “Identifying Bayesian Mixture Models,” mc-stan.org. \href{https://mc-stan.org/users/documentation/case-studies/identifying_mixture_models}{[Online]}.}. To avoid these issues, apart from the numerically robust formulation of the likelihoods in equations~\eqref{eq:moe-likelihood} and~\eqref{eq:cocoafuse-likelihood} proposed in Appendix~\ref{app:numerical-robustness}, we offer two implementations where the experts' symmetry is \textit{broken}, to prevent chains from swapping the experts (i.e., to contain the ``label switching'' problem).
	Specifically, we constrain the experts to be ``ordered'', in the sense that either:
	\begin{enumerate}[nosep]
		\item They have monotonically non-decreasing \textit{bias} coefficients;
		\item They have monotonically non-decreasing \textit{standard errors}.
	\end{enumerate}
	Another aspect that might affect the identifiability of the models is related to the gate's parameters, as introduced in equation~\eqref{eq:gate}. Indeed,
	the softmax function is invariant to translations of its inputs by identical offsets in each coordinate (i.e., by vectors $\bm{k} = [k, \dots, k]^\top$ with $k\in\mathbb{R}$). While we never rely on improper priors on the gate's parameters, this feature of the softmax function could lead to some pathological behaviors, which we address by setting the last column of $\bm{\theta}_\mathcal{G}$ to zero. In this way, the last expert always acts as a ``reference'' for the other inputs of the softmax.

	\subsection{Prior Elicitation}\label{ssec:prior-elicitation}
		When crafting reasonable priors (as the ones discussed in Appendix~\ref{app:choice-of-distr}) is impractical by hand,
		we follow the approach discussed in~\citet{2017wittenDataMiningPractical} to select prior hyperparameters via Marginal Likelihood Maximization.
		Since the (logarithm of the) Marginal Likelihood $L(\bm{\lambda}; \bm{y}, \bm{X})$ for the prior hyperparameters $\bm{\lambda}$ is intractable analytically, we consider as surrogate the Expected Log-Likelihood over the training dataset $\mathcal{D} = \left(\bm{X}, \bm{y}\right)$, defined as
		\begin{equation}\label{eq:expected-log-likelihood}
			\textup{ELL}(\bm{\lambda}) = \mathbb{E}_{\bm{\theta}\sim\textup{post}_{\bm{\lambda}}}\left[ \ln  L(\bm{\lambda};\bm{\theta}, \bm{X}, \bm{y}) \right] = \textup{ELL}_{\bm{\lambda}}(\bm{\theta}) + \mathbb{E}_{\bm{\theta}\sim\textup{post}_{\bm{\lambda}}}\left[ \ln  p(\bm{\theta} | \bm{\lambda}) \right]
		\end{equation}
		where the expectation is computed on the posterior density $\textup{post}_{\bm{\lambda}}$ given $\mathcal{D}$ for the set of priors indexed by the priors hyperparameters $\bm{\lambda}$. The following proposition explains why~\eqref{eq:expected-log-likelihood} is relevant to the maximization of $L(\bm{\lambda}; \bm{y}, \bm{X})$.
		\begin{propositions}[from~\citet{2017wittenDataMiningPractical}]\label{prop:witten-datamining}
			The derivative of the expected log-likelihood $\textup{ELL}(\bm{\lambda})$ with respect to the parameters of the model equals the derivative of the log-marginal likelihood $\ln L(\bm{\lambda}; \bm{y}, \bm{X})$.
		\end{propositions}
		To mitigate the %aforementioned
		expert collapse problem (see Section~\ref{ssec:label-switching-problem} and Appendix~\ref{app:model-select}), we also propose to augment the objective as %the quantity
		\begin{equation}\label{eq:eb-loss}
			 \mathcal{L}(\bm{\lambda}) = \ln L(\bm{\lambda}; \bm{y}, \bm{X}) + \gamma \mathcal{H}(\bm{\lambda}),
		\end{equation}
		with
		\begin{equation}\label{eq:entropy}
			 \mathcal{H}(\bm{\lambda}) = \mathbb{E}_{\bm{\theta}\sim\textup{post}_{\bm{\lambda}}}\left[
			\ln H\left(\bar{\bm{\alpha}}(\bm{\theta})\right)
			\right],
		\end{equation}
		where $H\left(\bm{\alpha}\right) = -\sum_{m=1}^M\alpha_m \log_2\left({\alpha_m}\right) / {\log_2(M)}$ is the (normalized) Shannon entropy computed on the averaged activations $\bar{\bm{\alpha}}(\bm{\theta}) = \frac{1}{N}\sum_{n=1}^N\bm{\alpha}(\bm{x}_n; \bm{\theta})$.
		The reasoning behind this choice is that we want to penalize solutions that underutilize one (or more) of the experts, as the logarithm of the Shannon entropy tends to $-\infty$ as components of $\bar{\bm{\alpha}}(\bm{\theta})$ go to zero.

		Jointly with Proposition~\ref{prop:witten-datamining}, the next proposition justifies a gradient ascent scheme on on~\eqref{eq:eb-loss} to maximize the Marginal Likelihood with respect to the hyperparameters, discussed more in detail in Appendix~\ref{app:empirical-bayes}.
		\begin{propositions}\label{prop:grad-entropy}
			The gradient of $\mathcal{H}(\bm{\lambda})$ in~\eqref{eq:entropy} with respect to $\bm{\lambda}$ is given by the expectation
			\begin{equation}
				 \nabla_{\bm{\lambda}} \mathcal{H}(\bm{\lambda}) = \mathbb{E}_{\bm{\theta}\sim\textup{post}_{\bm{\lambda}}}\left[
				\left(\ln H\left(\bar{\bm{\alpha}}(\bm{\theta})\right) - \mathcal{H}(\bm{\lambda}) \right) \nabla_{\bm{\lambda}} \ln p(\bm{\theta} | \bm{\lambda})
				\right].
			\end{equation}
		\end{propositions}

	\subsection{Model Selection}
		Model selection in mixtures of experts is notoriously a complex task, aggravated by the possibility of expert \enquote{collapse} (see~\citet{2023royerRevisitingSinglegatedMixtures}).
		Indeed, it can happen that one or more experts are excluded from the generation process as a result of the gate assigning them near-zero probabilities.
		Indeed, ``collapsed'' experts do not effectively contribute to the likelihoods~\eqref{eq:moe-likelihood} and~\eqref{eq:cocoafuse-likelihood}.
		Although a Bayesian treatment of this problem by assigning suitable priors as well as performing \textit{model post-processing} can help in mitigating improper model selection and overfitting, we find from experience that expert collapse is still possible and must be addressed explicitly at the model selection stage.
		We thus introduce a complexity-aware procedure for model selection, described in Algorithm~\ref{alg:pareto-model-selection} (Appendix~\ref{app:model-select}).
		This procediure was adopted to select the number of experts and gate complexity in the experiments of Section~\ref{sec:didactic-examples}.

	\section{Evaluation Metrics}\label{sec:eval-metrics}
	In our analyses, we are interested in modeling the entire conditional target distribution on the covariates. To assess this specific objective and the quality of models alongside the more usual ``average response fit'', we report a series of metrics which we coarsely group into the following categories.
	\paragraph{Posterior Predictive Density Fit.} To evaluate the \textit{generalization} properties of the fitted model (i.e., posterior sample) on the test data, we estimate via Monte Carlo sampling the Log Pointwise Predictive Density (LPPD), i.e.,
	\begin{equation}\label{eq:LPPD}
		 \textup{LPPD} = \sum_{i=1}^N\ln \left(\int p({y}_i | \bm{\theta}) p(\bm{\theta} | \mathcal{D})d\theta\right).
	\end{equation}
	For model evaluation purposes, we instead rely on the Pareto-smoothed importance sampling (PSIS) Leave-one-out (LOO) cross-validation introduced by~\citet{2017vehtariPracticalBayesianModel}, evaluated on the training set via the posterior sample.
	Since the PSIS-LOO is an estimator for the LPPD on unseen data, in both cases larger values indicate better fit.

	\paragraph{Credibility Coverage.} Together with the previous metrics, we report the expected 95\% Credible Interval Coverage (CIC), namely
	\begin{equation}\label{eq:CIC}
		 \textup{CIC}_{0.95} = \mathbb{P}_{\tilde{y}\sim f_{Y}}\left(\tilde{y}\in \textup{CI}_{0.95}\right)\approx \frac{1}{N}\sum_i\mathbb{I}_{{y}_i\in\textup{CI}_{0.95}},
	\end{equation}
	as well as the expected 95\% Credible Interval Length (CIL), i.e.,
	\begin{equation}\label{eq:CIL}
		 \textup{CIL}_{0.95} = \mathbb{E}_{\tilde{y}\sim f_{Y}}\left[\textup{length}(\textup{CI}_{0.95})\right]\approx \frac{1}{N}\sum_i\textup{length}(\textup{CI}_{0.95}({\bm{x}}_i)).
	\end{equation}
	Both measures are defined as means over the test set samples so that standard errors in the estimates can easily be attached to the estimates.
	Ideally, we expect the CIC to be \textit{very close} to the nominal value (e.g., 95\%). CICs being equal, we consider lower CIL to be better.

	\paragraph{Average Response Fit.} As a measure of the expected error on the predicted values under the square loss, we estimate the expected Mean Squared Error (eMSE)
	\begin{equation}\label{eq:eMSE}
		 \textup{eMSE} = \mathbb{E}_{\tilde{y}\sim f_{Y}}\left[\int \left(\tilde{y} - \hat{y}\right)^2 p(\hat{y} | \bm{\theta}) p(\bm{\theta} | \mathcal{D})d\theta\:\right]\approx \frac{1}{N}\frac{1}{S}\sum_i \sum_j \left({y}_i - \hat{y}^{(i)}_j\right)^2,
	\end{equation}
	where $\tilde{y}_i$ are the observations from the test set and $\hat{y}^{(i)}_j$ correspond to posterior predictive samples.
	Lower values of the eMSE indicate better fitting (on average) to the observed data.
	We also compute on all benchmark examples the Coefficient of Determination (R\textsuperscript{2}),
	\begin{equation}
		 R^2 = 1 - \sum_i \left(y_i - \bar{y}_i^S\right)^2 \Big/ \sum_i \left(y_i - \bar{y}\right)^2 \in [0,1],
	\end{equation}
	where $\bar{y}_i^S$ is a sampled estimate of the Posterior Predictive Mean at covariates $\bm{x}_i$ and $\bar{y}$ is the mean response over the data. Higher values of the Coefficient of Determination indicate better (Predictive Mean) fit.

	\begin{remarks}[Evaluation of Baseline Models]
		We use the Python package \texttt{pyro} (see~\citet{2019binghamPyroDeepUniversal}) to implement a small BNN (a single hidden layer with 5 hidden units) on each example.
		Inference on these models is performed via MCMC, thus all metrics are computed exactly alike MoE, BoE and CoCoAFusE.
		Instead, we adopt the Gaussian Process Regression (GPR) implementation in the \texttt{scikit-learn} package\footnote{Class~\href{https://scikit-learn.org/1.5/modules/generated/sklearn.gaussian_process.GaussianProcessRegressor}{\texttt{GaussianProcessRegressor}}.}, which does not return samples from the pointwise log-likelihoods under the posterior. Therefore, we cannot compute PSIS-LOO and LPPD (see Section~\ref{sec:eval-metrics}). We thus compare CoCoAFusE and the other models to the GPR in terms of coverage and average fit metrics.
	\end{remarks}

	\section{Experiments}\label{sec:experiments}

	In this section we discuss some realistic applications of our methodology.
	We compare the CoCoAFuse architecture to the standard MoE and previously described BoE on a suite of benchmark datasets: a publicly available Wind Turbine dataset, the Motorcycle Crash data from~\citet{1985silvermanAspectsSplineSmoothing}, and a Power Distribution Board Load Dataset from~\citet{2021yeafiPDBElectricPower}.
	We only consider univariate outputs $y$ for clarity of representation.
	Every experiment includes BNNs and GPRs (see Remark~\ref{remark:benchmark-models}) in order to assess the advantages of our multi-model approach against popular black-box alternatives.
	More details on each benchmark (including the specific choice of prior distributions as well as MCMC diagnostics\footnote{In none of the benchmark examples the $\hat{R}$ criterion yielded satisfactory results for the BNN, indicating a possible failed convergence to the posterior. Nevertheless, we report the results in the following.}) can be found in Appendix~\ref{app:details-benchmarks}.
	The code is freely available on GitHub\footnote{{\url{https://github.com/aurelio-raffa/cocoafuse.git}}}.

\begin{figure}[t!]
	\subfigure[CoCoAFusE]{\includegraphics[width=0.48\linewidth, trim={0.25cm 0 0.4cm 0.25cm}, clip]{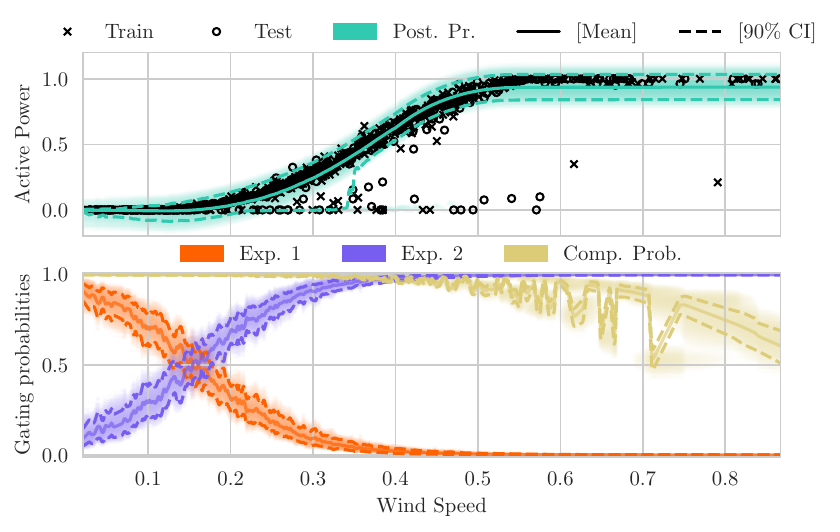}}
	\subfigure[MoE]{\includegraphics[width=0.48\linewidth, trim={0.25cm 0 0.4cm 0.25cm}, clip]{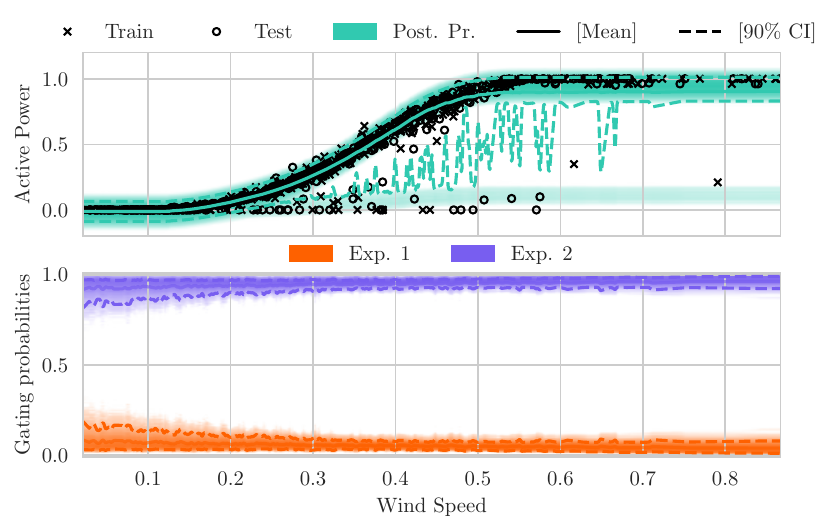}}\\
	\subfigure[BNN]{\includegraphics[width=0.48\linewidth, trim={0.26cm 2.75cm 0.4cm 0}, clip]{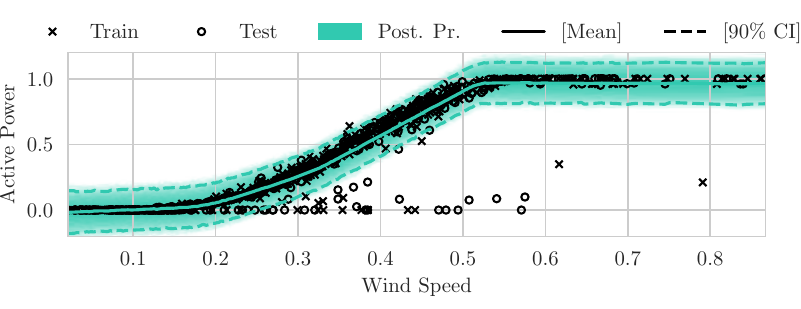}}
	\subfigure[GPR]{\includegraphics[width=0.48\linewidth, trim={0.26cm 2.75cm 0.4cm 0}, clip]{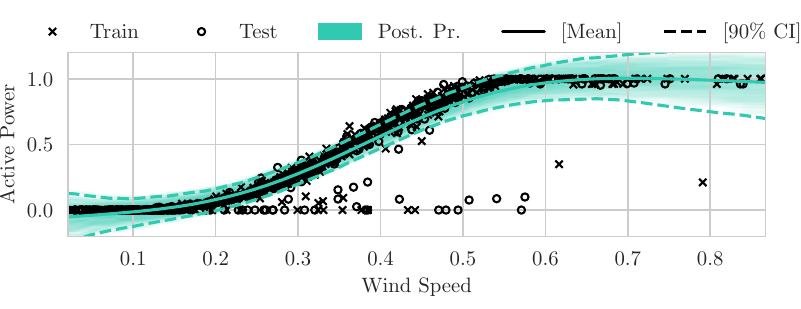}}
	\caption{Predictions for CoCoAFusE, MoE, BNN, and GPR on the Wind Turbine Data.}
	\label{fig:wind-turbine-posterior-predictive}
\end{figure}

	\begin{remarks}[Note on the Figures]
		Figures~\ref{fig:switch-posterior-predictive} though~\ref{fig:pdb-posterior-predictive} report an illustration of the posterior predictive density (top subplots), as well as the covariates-dependent expert allocation distribution (bottom subplots).
		More specifically, in the top subplots, we always display the data points (crosses for training observations vs. circles for test ones) against the predictive density (aqua). In the bottom subplots, we instead display each expert allocation conditional probability with a color specific to the corresponding expert.
		Densities are visualized via shades of the corresponding color, superimposed with continuous lines (indicating the posterior mean) and dashed lines (indicating the 90\% credible interval bounds).
		In Figures~\ref{fig:wind-turbine-posterior-predictive} though~\ref{fig:pdb-posterior-predictive}, we compare CoCoAFusE against the best (in terms of LPPD) among MoE and BoE, as well as the BNN and the GPR models from Remark~\ref{remark:benchmark-models}.
	\end{remarks}

	\begin{remarks}[Note on the Tables]
		The Tables report the PSIS-LOO computed on the Training Set, as well as the LPPD, 95\% Credible Interval Coverage (CIC), 95\% Credible Interval Length (CIL), eMSE, and R\textsuperscript{2} on the Test set.
		For the CIC, ideal scores correspond to the nominal 95\% value with small spread (i.e, standard errors), whereas for the CIL smaller values are better conditionally on the CIC fulfilling the aforementioned condition.
		Notice that for the Gaussian Process Regression (GPR), acting as baseline on the Benchmark Datasets, the adopted implementation prevented us from estimating PSIS-LOO and LPPD.

	\end{remarks}

	\subsection{Wind Turbine}
	As a first use case, we consider a dataset comprising $1000$ SCADA observations from a wind turbine located in the Yalova wind farm (offshore western Turkey). The dataset features occasional gaps and anomalous measurements due to turbine malfunctions, maintenance activities, or wind speeds falling below the cut-in threshold. This feature makes this dataset especially interesting to assess CoCoAFusE's capability to model the turbine's possible failure mode through a specific \textit{expert}. Hence, our goal with this example is to understand which among the modeling architectures best achieves a simultaneous description of the ``normal'' behavior of the turbine, which is not fully encompassed by the provided (static) theoretical curve, as well as anomalous states.
	We set out to do so by encoding our information on the various operating modes of the system through the priors on $M = 2$ experts (see Appendix~\ref{app:details-benchmarks}). Priors are identical MoE, BoE, and CoCoAFusE models for the common parts of the architecture.
	The MoE, BoE, CoCoAFusE, and GPR alternatives are applied to a sample of 500 randomly selected observations from the Wind Turbine dataset and evaluated on an additional 500, obtaining the indices reported in Table~\ref{tab:wind-turbine-scores}.
	\begin{table}[tb!]
		\caption{Figures of Merit (Best in bold \textpm\ Std. Error) on the Wind Turbine Example.}
		\label{tab:wind-turbine-scores}
		\centering
		\small
		\begin{tabular}{lcccccc}
			% table generated automatically on date 2025-01-31
			\toprule
			& Train & \multicolumn{5}{c}{Test} \\

			\cmidrule(lr){2-2}\cmidrule(lr){3-7}
			& $\uparrow$ LOO \footnotesize{[\texttimes10\textsuperscript{2}]}
			& $\uparrow$ LPPD \footnotesize{[\texttimes10\textsuperscript{2}]}
			& $_\uparrow^\downarrow$ 95\% CIC
			& $\downarrow$ 95\% CIL
			& $\downarrow$ eMSE \footnotesize{[\texttimes10\textsuperscript{-2}]}
			& $\uparrow$ R\textsuperscript{2} [\%] \\

			\midrule

			% data extracted automatically on date 2025-01-31 from path outputs/wind_turbine/moe/inference/evaluation.csv
			MoE &
			7.56\textpm0.39 &
			7.79\textpm0.32 &
			\textbf{0.96}\textpm0.01 &
			0.44\textpm0.01 &
			2.76\textpm0.40 &
			88.10\\

			% data extracted automatically on date 2025-01-31 from path outputs/wind_turbine/boe/inference/evaluation.csv
			BoE &
			8.71\textpm0.48 &
			\textbf{8.49}\textpm0.61 &
			0.93\textpm0.01 &
			0.29\textpm0.01 &
			2.51\textpm0.42 &
			87.87\\

			% data extracted automatically on date 2025-01-31 from path outputs/wind_turbine/cocoafuse/inference/evaluation.csv
			Co. &
			\textbf{9.47}\textpm0.51 &
			6.39\textpm2.03 &
			0.97\textpm0.01 &
			0.27\textpm0.01 &
			2.22\textpm0.42 &
			87.66\\

			% data extracted automatically on date 2025-02-18 from path outputs/wind_turbine/bnn/inference/evaluation.csv
			BNN &
			4.68\textpm0.67 &
			3.10\textpm1.06 &
			\textbf{0.96}\textpm0.01 &
			0.37\textpm0.00 &
			2.46\textpm0.41 &
			88.21\\

			% data extracted automatically on date 2025-01-31 from path outputs/wind_turbine/gpr/inference/evaluation.csv
			GPR &
			\textemdash &
			\textemdash &
			0.91\textpm0.01 &
			\textbf{0.24}\textpm0.00 &
			\textbf{1.96}\textpm0.37 &
			\textbf{88.34}\\

			\bottomrule
		\end{tabular}
	\end{table}
	In this example, the CoCoAFusE model is surpasses by the MoE and BoE in terms of LPPD (although it has a larger variance in the estimate), and by the MoE in terms of CIC, while at the same time it exhibit shorter intervals with a larger coverage in both cases and better eMSE than either.
	The BoE fails to capture ``anomalous'' observations (null active power), whereas CoCoAFusE and MoE successfully rely on mixing to describe them.
	The same can be said for the GPR and BNN models, both of which cannot capture multimodalities conditional to the same set of covariates. This is testified by the fact that the GPR model results in the worst coverage among tested models, despite the best average response fit, and by the fact that the BNN exhibits excessively large credible intervals.
	This suggests the perhaps obvious consideration that, in practical applications, achieving a good average response fit and a good density fit can point to different models.
	Simultaneously, we can notice how the MoE's predictions for the anomalous observations are shifted towards some outliers at high wind speeds, which does not happen with CoCoAFusE.

\begin{table}[tb!]
	\caption{Figures of Merit (Best in bold \textpm\ Std. Error) on the Motorcycle Crash Example.}
	\label{tab:motorcycle-scores}
	\centering
	\small
	\begin{tabular}{lcccccc}
		% table generated automatically on date 2025-01-31
		\toprule
		& Train & \multicolumn{5}{c}{Test} \\

		\cmidrule(lr){2-2}\cmidrule(lr){3-7}
		& $\uparrow$ LOO
		& $\uparrow$ LPPD
		& $_\uparrow^\downarrow$ 95\% CIC
		& $\downarrow$ 95\% CIL
		& $\downarrow$ eMSE [\texttimes10\textsuperscript{-2}]
		& $\uparrow$ R\textsuperscript{2} [\%] \\

		\midrule

		% data extracted automatically on date 2025-01-31 from path outputs/motorcycle/moe/inference/evaluation.csv
		MoE &
		46.78\textpm12.35 &
		\textbf{28.69}\textpm9.09 &
		0.98\textpm0.02 &
		0.80\textpm0.04 &
		8.51\textpm0.99 &
		\textbf{63.72}\\

		% data extracted automatically on date 2025-01-31 from path outputs/motorcycle/boe/inference/evaluation.csv
		BoE &
		\textbf{53.36}\textpm12.02 &
		25.36\textpm9.47 &
		0.90\textpm0.04 &
		\textbf{0.68}\textpm0.05 &
		8.10\textpm0.94 &
		58.93\\

		% data extracted automatically on date 2025-01-31 from path outputs/motorcycle/cocoafuse/inference/evaluation.csv
		CoCoA. &
		53.17\textpm12.10 &
		20.62\textpm10.98 &
		\textbf{0.95}\textpm0.03 &
		0.70\textpm0.05 &
		8.07\textpm0.94 &
		59.54\\

		% data extracted automatically on date 2025-02-18 from path outputs/motorcycle/bnn/inference/evaluation.csv
		BNN &
		4.29\textpm5.17 &
		10.95\textpm4.00 &
		0.98\textpm0.02 &
		0.93\textpm0.00 &
		9.42\textpm0.76 &
		62.53\\

		% data extracted automatically on date 2025-01-31 from path outputs/motorcycle/gpr/inference/evaluation.csv
		GPR &
		\textemdash &
		\textemdash &
		0.86\textpm0.05 &
		0.70\textpm0.03 &
		\textbf{7.78}\textpm0.84 &
		58.32\\

		\bottomrule
	\end{tabular}
\end{table}

	\subsection{Motorcycle Crash}
\begin{figure}[t!]
		\subfigure[CoCoAFusE]{\includegraphics[width=0.48\linewidth, trim={0.25cm 0 0.4cm 0.25cm}, clip]{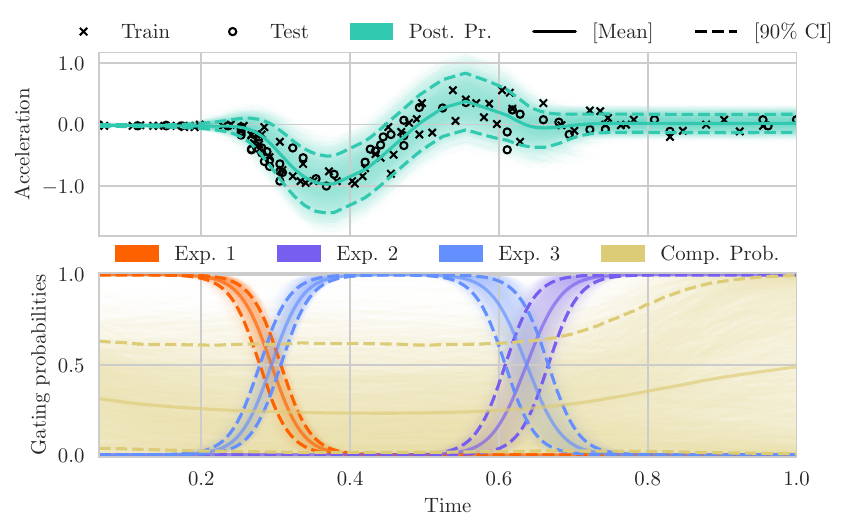}}
		\subfigure[MoE]{\includegraphics[width=0.48\linewidth, trim={0.25cm 0 0.4cm 0.25cm}, clip]{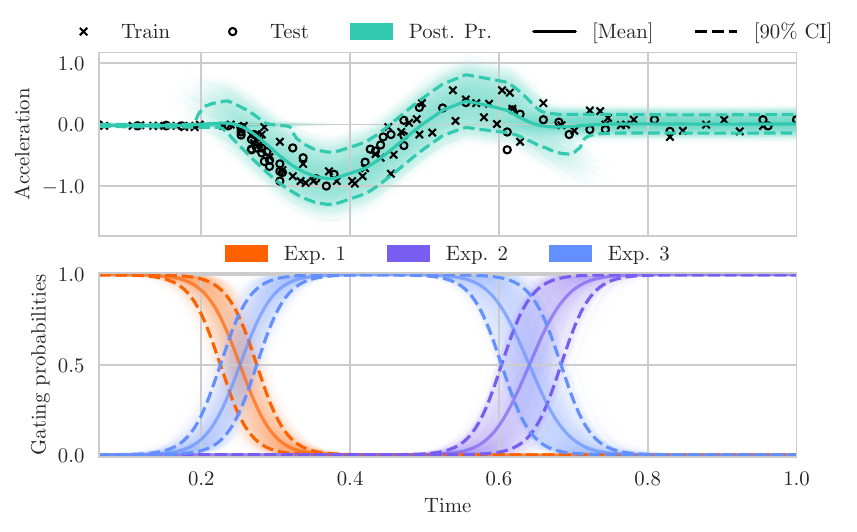}}\\
		\subfigure[BNN]{\includegraphics[width=0.48\linewidth, trim={0.26cm 2.5cm 0.4cm 0}, clip]{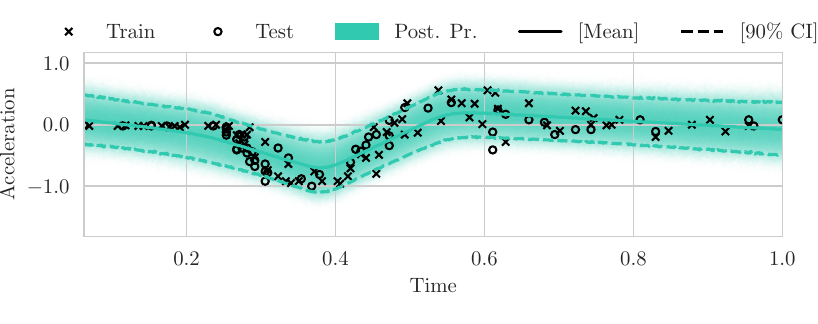}}
		\subfigure[GPR]{\includegraphics[width=0.48\linewidth, trim={0.26cm 2.5cm 0.4cm 0}, clip]{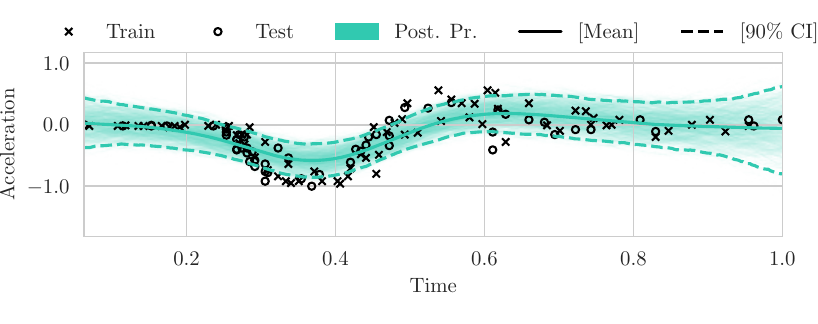}}
		\caption{Predictions for CoCoAFusE, MoE, BNN, and GPR on the Motorcycle Crash.}
		\label{fig:motorcycle-posterior-predictive}
	\end{figure}

	The Motorcycle Crash data, first analyzed in~\citet{1985silvermanAspectsSplineSmoothing}, has been used in several works concerning mixtures of experts (see, e.g.,~\citet{2001rasmussenInfiniteMixturesGaussian, 2005meedsAlternativeInfiniteMixture, 2012yukselTwentyYearsMixture}). Thus, we include it as a valid proving ground for CoCoAFusE. Of the 133 data points available, 75 observations are used to construct the training set, while the remaining 58 constitute our test set. By fixing the number of experts to $M=3$, our goal with this case study is to assess what can be the advantages of using CoCoAFuse over a classical MoE. % when informed priors are used.
	From visual inspection, we can easily formulate informative priors concerning the locations of the experts along the time coordinate.
	As seen in \figurename{~\ref{fig:motorcycle-posterior-predictive}}, CoCoAFusE results in low competition probabilities, which underscores that -- in this example -- relying on experts' \textit{blending}-only provides better predictive densities.
	This observation is corroborated by the metrics in \tablename{~\ref{tab:motorcycle-scores}}.
	Although the differences in metrics are not statistically significant, we can observe better coverage as well as shorter credible intervals for CoCoAFusE compared to MoE.
	Moreover, qualitative aspects set the MoE and BoE/CoCoAFusE apart. Through collaborative behavior, we witness better transitions from one expert ``domain'' to the next.
	This is especially visible in the areas where experts have comparable weights, in which the MoE
	exhibits large CIs for the acceleration,
	while the CoCoAFusE and BoE produce a funnel-like shape consistent with a smooth change in variance.
	By constrast, we can observe that the GPR results in an insufficient coverage of the observations, whereas the BNN yields inflated credible intervals.

\subsection{Power Distribution Board}

	\begin{table}[t!]
		\caption{Figures of Merit (Best in bold \textpm\ Std. Error) on the Power Distribution Board.}
		\label{tab:pdb-scores}
		\centering
		\small
		\begin{tabular}{lcccccc}
			% table generated automatically on date 2025-01-31
			\toprule
			& Train & \multicolumn{5}{c}{Test} \\

			\cmidrule(lr){2-2}\cmidrule(lr){3-7}
			& $\uparrow$ LOO [\texttimes10\textsuperscript{2}]
			& $\uparrow$ LPPD [\texttimes10\textsuperscript{2}]
			& $_\uparrow^\downarrow$ 95\% CIC
			& $\downarrow$ 95\% CIL
			& $\downarrow$ eMSE
			& $\uparrow$ R\textsuperscript{2} [\%] \\

			\midrule

			% data extracted automatically on date 2025-01-31 from path outputs/power_distribution_board/moe/inference/evaluation.csv
			MoE &
			-5.66\textpm0.24 &
			-1.09\textpm0.11 &
			0.96\textpm0.01 &
			1.69\textpm0.02 &
			0.32\textpm0.02 &
			87.34\\

			% data extracted automatically on date 2025-01-31 from path outputs/power_distribution_board/boe/inference/evaluation.csv
			BoE &
			-2.43\textpm0.25 &
			-0.53\textpm0.10 &
			\textbf{0.95}\textpm0.01 &
			1.18\textpm0.01 &
			0.19\textpm0.01 &
			89.90\\

			% data extracted automatically on date 2025-01-31 from path outputs/power_distribution_board/cocoafuse/inference/evaluation.csv
			CoCoA. &
			-2.09\textpm0.25 &
			-0.52\textpm0.11 &
			0.94\textpm0.01 &
			1.13\textpm0.01 &
			0.18\textpm0.01 &
			90.41\\

			% data extracted automatically on date 2025-02-18 from path outputs/power_distribution_board/bnn/inference/evaluation.csv
			BNN &
			\textbf{-0.37}\textpm0.33 &
			\textbf{-0.21}\textpm0.15 &
			\textbf{0.95}\textpm0.01 &
			0.98\textpm0.00 &
			\textbf{0.13}\textpm0.01 &
			\textbf{92.62}\\

			% data extracted automatically on date 2025-01-31 from path outputs/power_distribution_board/gpr/inference/evaluation.csv
			GPR &
			\textemdash &
			\textemdash &
			0.83\textpm0.02 &
			\textbf{0.80}\textpm0.01 &
			\textbf{0.13}\textpm0.01 &
			91.21\\

			\bottomrule
		\end{tabular}
	\end{table}

\begin{figure}[t!]
		\subfigure[CoCoAFusE]{\includegraphics[width=0.48\linewidth, trim={0.25cm 0 0.25cm 0.25cm}, clip]{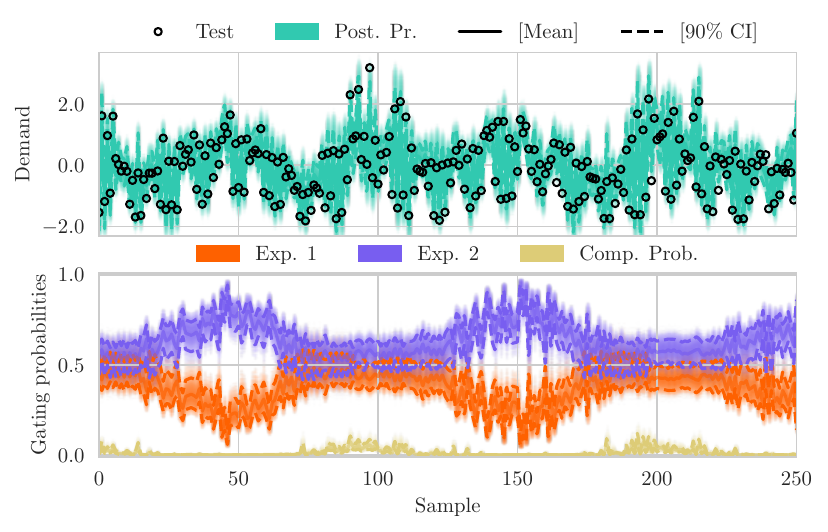}}
		\subfigure[BoE]{\includegraphics[width=0.48\linewidth, trim={0.25cm 0 0.25cm 0.25cm}, clip]{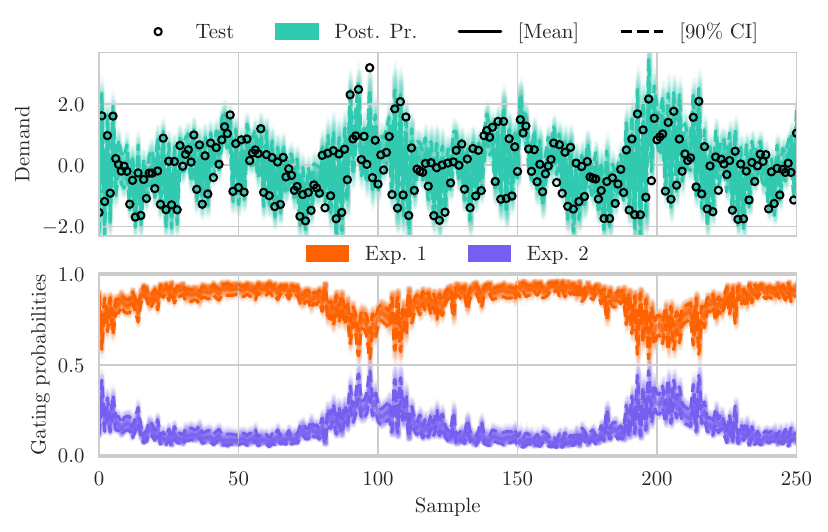}}\\
		\subfigure[BNN]{\includegraphics[width=0.48\linewidth, trim={0.26cm 2.5cm 0.25cm 0}, clip]{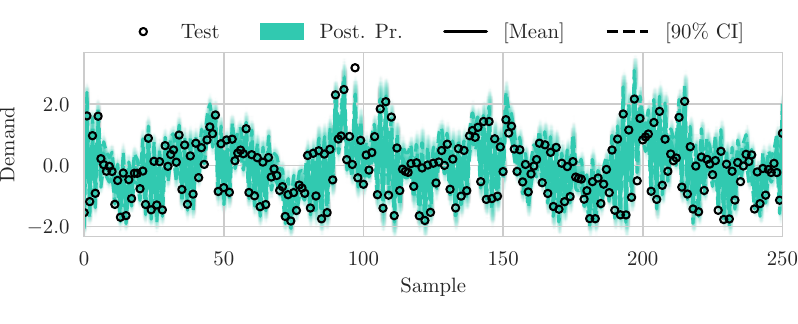}}
		\subfigure[GPR]{\includegraphics[width=0.48\linewidth, trim={0.26cm 2.5cm 0.25cm 0}, clip]{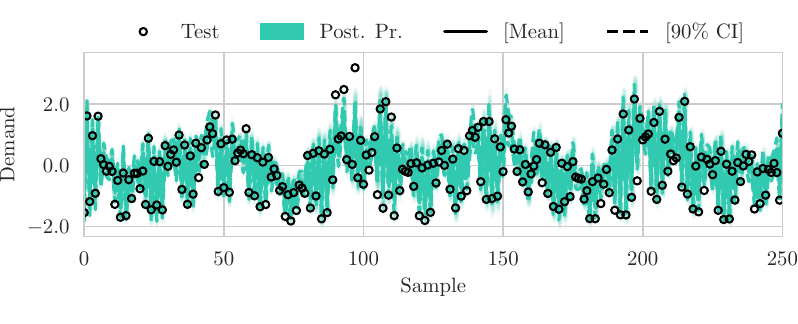}}
		\caption{CoCoAFusE, BoE, BNN, and GPR on the Power Distribution Board Data.}
		\label{fig:pdb-posterior-predictive}
	\end{figure}

	To complete our suite of examples, we introduce a case study where it is difficult to formulate informative priors by visual inspection.
	The data we adopt here is a time history of the aggregated power demand of a distribution network, collected from readings of a Power Distribution Board (PBD) over a span of 11 years with an hourly frequency.
	Other than the time information, the dataset contains the air temperature as an exogenous signal.
	With the large set of features we define (see Appendix~\ref{app:details-benchmarks}), crafting convenient priors by hand is a prohibitive task.
	Therefore, we tune the prior hyperparameters via the Empirical Bayes approach described in Section~\ref{ssec:prior-elicitation} and Appendix~\ref{app:empirical-bayes}.
	From the available dataset, we draw 1000 samples randomly from the first 80\% of the data and 250 equispaced samples from the last 20\%.
	The EB procedure yields a model that is extremely skewed towards collaboration, as visible in \figurename{~\ref{fig:pdb-posterior-predictive}}.
	On the other hand, imposing the competition (via MoE with identical priors) leads to a significantly worse metrics.
	The GPR still exhibits better average fit than MoE, BoE, and CoCoAFusE but the worst of the coverage metrics, with several extreme data points falling outside of the GPR's credible bounds but inside CoCoAFusE's.
	In this example, however, the BNN surpasses all other models on almost all metrics, suggesting that the linear experts might be insufficient to accurately model the intricate input-output dependency in the data.
	However, this model also features the worst convergence diagnostics (see Appendix~\ref{app:convergence-diagnostics}), which highlights the difficulty of working with heavily nonlinear models within a Bayesian framework.

	\section{Discussion and Conclusions}
	In this work, we have introduced a novel Bayesian architecture, the CoCoAFusE, rooted in the same philosophy of the Mixture of Experts. With the goal of tackling density fitting problems characterized by a multiplicity of \textit{data regimes} and featuring either abrupt stochastic switches, smooth transitions, or intermediate scenarios, our approach is guided by considerations about parsimony as well as interpretability of the representations.
	Through a set of examples, we have showcased the limitations of finite MoEs (and BoEs) and demonstrated how CoCoAFusE helps in overcoming them.
	The key take-away is that we can expect from CoCoAFusE a performance that is al least comparable with the best among the MoE and BoE, which are special cases of CoCoAFusE.
	Furthermore, we have compared CoCoAFusE with GPR and BNN models, two among the most popular probabilistic modeling approaches in Machine Learning.
	While we are certain that more sophisticated approaches can significantly improve the GPR and BNN results, we advance a few important considerations regarding their limitations, which CoCoAFusE attempts to address:
	\begin{enumerate}[nosep]
		\item GPR and BNN are fully black-box models, thus not easily interpretable and it is nontrivial to design informative priors for either of them;
		\item BNNs are complex, with entire hierarchies (i.e., layers) of latent variables, resulting in poor convergence and high computational costs of MCMCs;
		\item Both GPR and BNN  are flexible in terms of the response's \textit{dependency} but not necessarily of its \textit{conditional density}, which might result in good average response albeit poor density fit (see, for example, the results in~\tablename{~\ref{tab:wind-turbine-scores}} and~\tablename{~\ref{tab:pdb-scores}}).
	\end{enumerate}

	\paragraph{Limitations.} The current CoCoAFusE framework comes with some limitations.
	Notably, the \emph{computational overhead} of performing inference is high, limiting both the overall model complexity attainable and applicability to large datasets.
	We aim to explore better-tailored implementations and sampling schemes, as well as approximate inference techniques.

	\acknowledgement{The work of {M. Tanelli} was partially supported by the following: {PRIN project TECHIE: “A control and network-based approach for fostering the adoption of new technologies in the ecological transition” Cod. 2022KPHA24 CUP: D53D23001320006; MOST – Sustainable Mobility National Research Center that received funding from the European Union NextGenerationEU (PIANO NAZIONALE DI RIPRESA E RESILIENZA (PNRR) – MISSIONE 4 COMPONENTE 2, INVESTIMENTO 1.4-D.D. 1033 17/06/2022, CN00000023), Spoke 5 “Light Vehicle and Active Mobility”.}
	The authors declare no conflict of interest.}
	\bibliographystyle{unsrtnat}
	\bibliography{cocoafuse_arxiv_2025.bib}

\begin{thebibliography}{48}
\providecommand{\natexlab}[1]{#1}
\providecommand{\url}[1]{\texttt{#1}}
\expandafter\ifx\csname urlstyle\endcsname\relax
  \providecommand{\doi}[1]{doi: #1}\else
  \providecommand{\doi}{doi: \begingroup \urlstyle{rm}\Url}\fi

\bibitem[Zheng et~al.(2019)Zheng, Yuan, Zhu, Lin, Cheng, Shi, and
  Ye]{2019zhengSelfSupervisedMixtureofExpertsUncertainty}
Zhuobin Zheng, Chun Yuan, Xinrui Zhu, Zhihui Lin, Yangyang Cheng, Cheng Shi,
  and Jiahui Ye.
\newblock Self-{{Supervised Mixture-of-Experts}} by {{Uncertainty Estimation}}.
\newblock \emph{Proceedings of the AAAI Conference on Artificial Intelligence},
  33\penalty0 (01):\penalty0 5933--5940, July 2019.
\newblock ISSN 2374-3468, 2159-5399.
\newblock \doi{10.1609/aaai.v33i01.33015933}.

\bibitem[Hazimeh et~al.(2021)Hazimeh, Zhao, Chowdhery, Sathiamoorthy, Chen,
  Mazumder, Hong, and Chi]{2021hazimehDSelectkDifferentiableSelection}
Hussein Hazimeh, Zhe Zhao, Aakanksha Chowdhery, Maheswaran Sathiamoorthy, Yihua
  Chen, Rahul Mazumder, Lichan Hong, and Ed~Chi.
\newblock {{DSelect-k}}: {{Differentiable Selection}} in the {{Mixture}} of
  {{Experts}} with {{Applications}} to {{Multi-Task Learning}}.
\newblock In \emph{Advances in {{Neural Information Processing Systems}}},
  volume~34, pages 29335--29347. Curran Associates, Inc., 2021.

\bibitem[Ponti et~al.(2022)Ponti, Sordoni, Bengio, and
  Reddy]{2022pontiCombiningModularSkills}
Edoardo~M. Ponti, Alessandro Sordoni, Yoshua Bengio, and Siva Reddy.
\newblock Combining {{Modular Skills}} in {{Multitask Learning}}, March 2022.

\bibitem[Fedus et~al.(2022)Fedus, Zoph, and
  Shazeer]{2022fedusSwitchTransformersScaling}
William Fedus, Barret Zoph, and Noam Shazeer.
\newblock Switch {{Transformers}}: {{Scaling}} to {{Trillion Parameter Models}}
  with {{Simple}} and {{Efficient Sparsity}}, June 2022.

\bibitem[Zhao et~al.(2023)Zhao, Zhao, Shi, Kuang, and
  Liu]{2023zhaoCollaborativeMixtureofExpertsModel}
Jian Zhao, Zisong Zhao, Lijuan Shi, Zhejun Kuang, and Yazhou Liu.
\newblock Collaborative {{Mixture-of-Experts Model}} for {{Multi-Domain Fake
  News Detection}}.
\newblock \emph{Electronics}, 12\penalty0 (16):\penalty0 3440, January 2023.
\newblock ISSN 2079-9292.
\newblock \doi{10.3390/electronics12163440}.

\bibitem[Abdar et~al.(2021)Abdar, Pourpanah, Hussain, Rezazadegan, Liu,
  Ghavamzadeh, Fieguth, Cao, Khosravi, Acharya, Makarenkov, and
  Nahavandi]{2021abdarReviewUncertaintyQuantification}
Moloud Abdar, Farhad Pourpanah, Sadiq Hussain, Dana Rezazadegan, Li~Liu,
  Mohammad Ghavamzadeh, Paul Fieguth, Xiaochun Cao, Abbas Khosravi, U.~Rajendra
  Acharya, Vladimir Makarenkov, and Saeid Nahavandi.
\newblock A review of uncertainty quantification in deep learning:
  {{Techniques}}, applications and challenges.
\newblock \emph{Information Fusion}, 76:\penalty0 243--297, December 2021.
\newblock ISSN 1566-2535.
\newblock \doi{10.1016/j.inffus.2021.05.008}.

\bibitem[Burkart and Huber(2021)]{2021burkartSurveyExplainabilitySupervised}
Nadia Burkart and Marco~F Huber.
\newblock A {{Survey}} on the {{Explainability}} of {{Supervised Machine
  Learning}}.
\newblock \emph{Journal of Artificial Intelligence Research}, 70, January 2021.
\newblock \doi{10.1613/jair.1.12228}.

\bibitem[Tambon et~al.(2022)Tambon, Laberge, An, Nikanjam, Mindom, Pequignot,
  Khomh, Antoniol, Merlo, and Laviolette]{2022tambonHowCertifyMachine}
Florian Tambon, Gabriel Laberge, Le~An, Amin Nikanjam, Paulina Stevia~Nouwou
  Mindom, Yann Pequignot, Foutse Khomh, Giulio Antoniol, Ettore Merlo, and
  Fran{\c c}ois Laviolette.
\newblock How to certify machine learning based safety-critical systems? {{A}}
  systematic literature review.
\newblock \emph{Automated Software Engineering}, 29\penalty0 (2):\penalty0 38,
  April 2022.
\newblock ISSN 1573-7535.
\newblock \doi{10.1007/s10515-022-00337-x}.

\bibitem[Draper(1995)]{1995draperAssessmentPropagationModel}
David Draper.
\newblock Assessment and {{Propagation}} of {{Model Uncertainty}}.
\newblock \emph{Journal of the Royal Statistical Society: Series B
  (Methodological)}, 57\penalty0 (1):\penalty0 45--70, 1995.
\newblock ISSN 2517-6161.
\newblock \doi{10.1111/j.2517-6161.1995.tb02015.x}.

\bibitem[Bernardo and Smith(2009)]{2009bernardoBayesianTheory}
Jos{\'e}~M. Bernardo and Adrian F.~M. Smith.
\newblock \emph{Bayesian {{Theory}}}.
\newblock John Wiley \& Sons, September 2009.
\newblock ISBN 978-0-470-31771-6.

\bibitem[Zyphur and Oswald(2013)]{2013zyphurBayesianEstimationInference}
Michael~J. Zyphur and Frederick~L. Oswald.
\newblock Bayesian {{Estimation}} and {{Inference}}.
\newblock \emph{Journal of Management}, 41\penalty0 (2), August 2013.
\newblock \doi{10.1177/0149206313501200}.

\bibitem[Gelman et~al.(2020)Gelman, Vehtari, Simpson, Margossian, Carpenter,
  Yao, Kennedy, Gabry, B{\"u}rkner, and Modr{\'a}k]{2020gelmanBayesianWorkflow}
Andrew Gelman, Aki Vehtari, Daniel Simpson, Charles~C. Margossian, Bob
  Carpenter, Yuling Yao, Lauren Kennedy, Jonah Gabry, Paul-Christian
  B{\"u}rkner, and Martin Modr{\'a}k.
\newblock Bayesian {{Workflow}}, November 2020.

\bibitem[Wade et~al.(2023)Wade, Inacio, and
  Petrone]{2023wadeBayesianDependentMixture}
Sara Wade, Vanda Inacio, and Sonia Petrone.
\newblock Bayesian dependent mixture models: {{A}} predictive comparison and
  survey, July 2023.

\bibitem[Luengo et~al.(2020)Luengo, Martino, Bugallo, Elvira, and
  S{\"a}rkk{\"a}]{2020luengoSurveyMonteCarlo}
David Luengo, Luca Martino, M{\'o}nica Bugallo, V{\'i}ctor Elvira, and Simo
  S{\"a}rkk{\"a}.
\newblock A survey of {{Monte Carlo}} methods for parameter estimation.
\newblock \emph{EURASIP Journal on Advances in Signal Processing},
  2020\penalty0 (1):\penalty0 25, May 2020.
\newblock ISSN 1687-6180.
\newblock \doi{10.1186/s13634-020-00675-6}.

\bibitem[Gunapati et~al.(2022)Gunapati, Jain, Srijith, and
  Desai]{2022gunapatiVariationalInferenceAlternative}
Geetakrishnasai Gunapati, Anirudh Jain, P.~K. Srijith, and Shantanu Desai.
\newblock Variational {{Inference}} as an alternative to {{MCMC}} for parameter
  estimation and model selection.
\newblock \emph{Publications of the Astronomical Society of Australia},
  39:\penalty0 e001, 2022.
\newblock ISSN 1323-3580, 1448-6083.
\newblock \doi{10.1017/pasa.2021.64}.

\bibitem[Ayd{\i}nhan et~al.(2024)Ayd{\i}nhan, Kolm, Mulvey, and
  Shu]{2024aydinhanIdentifyingPatternsFinancial}
Af{\c s}ar~Onat Ayd{\i}nhan, Petter~N. Kolm, John~M. Mulvey, and Yizhan Shu.
\newblock Identifying patterns in financial markets: Extending the statistical
  jump model for regime identification.
\newblock \emph{Annals of Operations Research}, May 2024.
\newblock ISSN 1572-9338.
\newblock \doi{10.1007/s10479-024-06035-z}.

\bibitem[Callaham et~al.(2021)Callaham, Koch, Brunton, Kutz, and
  Brunton]{2021callahamLearningDominantPhysical}
Jared~L. Callaham, James~V. Koch, Bingni~W. Brunton, J.~Nathan Kutz, and
  Steven~L. Brunton.
\newblock Learning dominant physical processes with data-driven balance models.
\newblock \emph{Nature Communications}, 12\penalty0 (1):\penalty0 1016,
  February 2021.
\newblock ISSN 2041-1723.
\newblock \doi{10.1038/s41467-021-21331-z}.

\bibitem[Lundbergh et~al.(2003)Lundbergh, Ter{\"a}svirta, and {van
  Dijk}]{2003lundberghTimeVaryingSmoothTransition}
Stefan Lundbergh, Timo Ter{\"a}svirta, and Dick {van Dijk}.
\newblock Time-{{Varying Smooth Transition Autoregressive Models}}.
\newblock \emph{Journal of Business \& Economic Statistics}, 21\penalty0
  (1):\penalty0 104--121, January 2003.
\newblock ISSN 0735-0015.
\newblock \doi{10.1198/073500102288618810}.

\bibitem[Iamsumang et~al.(2015)Iamsumang, Mosleh, and
  Modarres]{2015iamsumangHybridDBNMonitoring}
Chonlagarn Iamsumang, Ali Mosleh, and Mohammad Modarres.
\newblock Hybrid {{DBN}} monitoring and anomaly detection algorithms for
  on-line {{SHM}}.
\newblock In \emph{2015 {{Annual Reliability}} and {{Maintainability
  Symposium}} ({{RAMS}})}, pages 1--7, January 2015.
\newblock \doi{10.1109/RAMS.2015.7105184}.

\bibitem[Hu and Li(2020)]{2020huAnomalyDetectionRemaining}
Xiangzhi Hu and Ning Li.
\newblock Anomaly {{Detection}} and {{Remaining Useful Lifetime Estimation
  Based}} on {{Degradation State}} for {{Bearings}}.
\newblock In \emph{2020 39th {{Chinese Control Conference}} ({{CCC}})}, pages
  4216--4221, July 2020.
\newblock \doi{10.23919/CCC50068.2020.9188410}.

\bibitem[{van de Schoot} et~al.(2021){van de Schoot}, Depaoli, King, Kramer,
  M{\"a}rtens, Tadesse, Vannucci, Gelman, Veen, Willemsen, and
  Yau]{2021vandeschootBayesianStatisticsModellinga}
Rens {van de Schoot}, Sarah Depaoli, Ruth King, Bianca Kramer, Kaspar
  M{\"a}rtens, Mahlet~G. Tadesse, Marina Vannucci, Andrew Gelman, Duco Veen,
  Joukje Willemsen, and Christopher Yau.
\newblock Bayesian statistics and modelling.
\newblock \emph{Nature Reviews Methods Primers}, 1\penalty0 (1):\penalty0
  1--26, January 2021.
\newblock ISSN 2662-8449.
\newblock \doi{10.1038/s43586-020-00001-2}.

\bibitem[Rasmussen and Williams(2008)]{2008rasmussenGaussianProcessesMachine}
Carl~Edward Rasmussen and Christopher K.~I. Williams.
\newblock \emph{Gaussian Processes for Machine Learning}.
\newblock Adaptive Computation and Machine Learning. MIT Press, Cambridge,
  Mass., 3. print edition, 2008.
\newblock ISBN 978-0-262-18253-9.

\bibitem[Jospin et~al.(2022)Jospin, Laga, Boussaid, Buntine, and
  Bennamoun]{2022jospinHandsOnBayesianNeural}
Laurent~Valentin Jospin, Hamid Laga, Farid Boussaid, Wray Buntine, and Mohammed
  Bennamoun.
\newblock Hands-{{On Bayesian Neural Networks}}---{{A Tutorial}} for {{Deep
  Learning Users}}.
\newblock \emph{IEEE Computational Intelligence Magazine}, 17\penalty0
  (2):\penalty0 29--48, May 2022.
\newblock ISSN 1556-603X, 1556-6048.
\newblock \doi{10.1109/MCI.2022.3155327}.

\bibitem[Bursal(1996)]{1996bursalInterpolatingProbabilityDistributions}
Faruk~H. Bursal.
\newblock On interpolating between probability distributions.
\newblock \emph{Applied Mathematics and Computation}, 77\penalty0 (2):\penalty0
  213--244, July 1996.
\newblock ISSN 0096-3003.
\newblock \doi{10.1016/S0096-3003(95)00216-2}.

\bibitem[Am{\'e}ndola et~al.(2020)Am{\'e}ndola, Engstr{\"o}m, and
  Haase]{2020amendolaMaximumNumberModes}
Carlos Am{\'e}ndola, Alexander Engstr{\"o}m, and Christian Haase.
\newblock Maximum number of modes of {{Gaussian}} mixtures.
\newblock \emph{Information and Inference: A Journal of the IMA}, 9\penalty0
  (3):\penalty0 587--600, September 2020.
\newblock ISSN 2049-8772.
\newblock \doi{10.1093/imaiai/iaz013}.

\bibitem[Jacobs et~al.(1991)Jacobs, Jordan, Nowlan, and
  Hinton]{1991jacobsAdaptiveMixturesLocal}
Robert~A. Jacobs, Michael~I. Jordan, Steven~J. Nowlan, and Geoffrey~E. Hinton.
\newblock Adaptive {{Mixtures}} of {{Local Experts}}.
\newblock \emph{Neural Computation}, 3\penalty0 (1):\penalty0 79--87, March
  1991.
\newblock ISSN 0899-7667.
\newblock \doi{10.1162/neco.1991.3.1.79}.

\bibitem[Kanaujia and Metaxas(2006)]{2006kanaujiaLearningAmbiguitiesUsing}
Atul Kanaujia and Dimitris Metaxas.
\newblock Learning {{Ambiguities Using Bayesian Mixture}} of {{Experts}}.
\newblock In \emph{2006 18th {{IEEE International Conference}} on {{Tools}}
  with {{Artificial Intelligence}} ({{ICTAI}}'06)}, pages 436--440, Arlington,
  VA, November 2006. IEEE.
\newblock ISBN 978-0-7695-2728-4.
\newblock \doi{10.1109/ICTAI.2006.73}.

\bibitem[Yuksel et~al.(2012)Yuksel, Wilson, and
  Gader]{2012yukselTwentyYearsMixture}
Seniha~Esen Yuksel, Joseph~N. Wilson, and Paul~D. Gader.
\newblock Twenty {{Years}} of {{Mixture}} of {{Experts}}.
\newblock \emph{IEEE Transactions on Neural Networks and Learning Systems},
  23\penalty0 (8):\penalty0 1177--1193, August 2012.
\newblock ISSN 2162-2388.
\newblock \doi{10.1109/TNNLS.2012.2200299}.

\bibitem[Baldacchino et~al.(2017)Baldacchino, Worden, and
  Rowson]{2017baldacchinoRobustNonlinearSystem}
Tara Baldacchino, Keith Worden, and Jennifer Rowson.
\newblock Robust nonlinear system identification: {{Bayesian}} mixture of
  experts using the t-distribution.
\newblock \emph{Mechanical Systems and Signal Processing}, 85:\penalty0
  977--992, February 2017.
\newblock ISSN 0888-3270.
\newblock \doi{10.1016/j.ymssp.2016.08.045}.

\bibitem[Waterhouse et~al.(1995)Waterhouse, MacKay, and
  Robinson]{1995waterhouseBayesianMethodsMixtures}
Steve Waterhouse, David MacKay, and Anthony Robinson.
\newblock Bayesian {{Methods}} for {{Mixtures}} of {{Experts}}.
\newblock In \emph{Advances in {{Neural Information Processing Systems}}},
  volume~8. MIT Press, 1995.

\bibitem[Xia et~al.(2020)Xia, Richard~Hahn, and
  Gustafson]{2020xiaBayesianMixtureExperts}
Michelle Xia, P.~Richard~Hahn, and Paul Gustafson.
\newblock A {{Bayesian}} mixture of experts approach to covariate
  misclassification.
\newblock \emph{Canadian Journal of Statistics}, 48\penalty0 (4):\penalty0
  731--750, 2020.
\newblock ISSN 1708-945X.
\newblock \doi{10.1002/cjs.11560}.

\bibitem[Bishop and Svensen(2012)]{2012bishopBayesianHierarchicalMixtures}
Christopher~M. Bishop and Markus Svensen.
\newblock Bayesian {{Hierarchical Mixtures}} of {{Experts}}, October 2012.

\bibitem[Zhang et~al.(2022)Zhang, Bokrantz, and
  Olsson]{2022zhangSimilaritybasedBayesianMixtureofexperts}
Tianfang Zhang, Rasmus Bokrantz, and Jimmy Olsson.
\newblock A similarity-based {{Bayesian}} mixture-of-experts model, August
  2022.

\bibitem[Richardson and Green(1997)]{1997richardsonBayesianAnalysisMixtures}
{\relax Sylvia}.~Richardson and Peter~J. Green.
\newblock On {{Bayesian Analysis}} of {{Mixtures}} with an {{Unknown Number}}
  of {{Components}} (with discussion).
\newblock \emph{Journal of the Royal Statistical Society: Series B
  (Methodological)}, 59\penalty0 (4):\penalty0 731--792, November 1997.
\newblock ISSN 0035-9246.
\newblock \doi{10.1111/1467-9868.00095}.

\bibitem[Rasmussen and Ghahramani(2001)]{2001rasmussenInfiniteMixturesGaussian}
Carl Rasmussen and Zoubin Ghahramani.
\newblock Infinite {{Mixtures}} of {{Gaussian Process Experts}}.
\newblock In \emph{Advances in {{Neural Information Processing Systems}}},
  volume~14. MIT Press, 2001.

\bibitem[Meeds and Osindero(2005)]{2005meedsAlternativeInfiniteMixture}
Edward Meeds and Simon Osindero.
\newblock An {{Alternative Infinite Mixture Of Gaussian Process Experts}}.
\newblock In \emph{Advances in {{Neural Information Processing Systems}}},
  volume~18. MIT Press, 2005.

\bibitem[Casau et~al.(2020)Casau, Cunha, Sanfelice, and
  Silvestre]{2020casauHybridControlRobust}
Pedro Casau, Rita Cunha, Ricardo~G. Sanfelice, and Carlos Silvestre.
\newblock Hybrid {{Control}} for {{Robust}} and {{Global Tracking}} on {{Smooth
  Manifolds}}.
\newblock \emph{IEEE Transactions on Automatic Control}, 65\penalty0
  (5):\penalty0 1870--1885, May 2020.
\newblock ISSN 1558-2523.
\newblock \doi{10.1109/TAC.2019.2927708}.

\bibitem[Moradvandi et~al.(2023)Moradvandi, Lindeboom, Abraham, and
  De~Schutter]{2023moradvandiModelsMethodsHybrid}
Ali Moradvandi, Ralph E.~F. Lindeboom, Edo Abraham, and Bart De~Schutter.
\newblock Models and methods for hybrid system identification: A systematic
  survey{\emph{*}}.
\newblock \emph{IFAC-PapersOnLine}, 56\penalty0 (2):\penalty0 95--107, January
  2023.
\newblock ISSN 2405-8963.
\newblock \doi{10.1016/j.ifacol.2023.10.1553}.

\bibitem[Hoffman and Gelman(2014)]{2014hoffmanNoUTurnSamplerAdaptively}
Matthew~D Hoffman and Andrew Gelman.
\newblock The {{No-U-Turn Sampler}}: {{Adaptively Setting Path Lengths}} in
  {{Hamiltonian Monte Carlo}}.
\newblock \emph{Journal of Machine Learning Research}, 15, April 2014.

\bibitem[Witten et~al.(2017)Witten, Frank, Hall, and
  Pal]{2017wittenDataMiningPractical}
Ian~H. Witten, Eibe Frank, Mark~A. Hall, and Christopher~J. Pal.
\newblock \emph{Data Mining: Practical Machine Learning Tools and Techniques}.
\newblock Elsevier, Morgan Kaufmann, Amsterdam Boston Heidelberg London New
  York Oxford Paris San Diego San Francisco Singapore Sydney Tokyo, fourth
  edition edition, 2017.
\newblock ISBN 978-0-12-804291-5.

\bibitem[Royer et~al.(2023)Royer, Karmanov, Skliar, Bejnordi, and
  Blankevoort]{2023royerRevisitingSinglegatedMixtures}
Amelie Royer, Ilia Karmanov, Andrii Skliar, Babak~Ehteshami Bejnordi, and
  Tijmen Blankevoort.
\newblock Revisiting {{Single-gated Mixtures}} of {{Experts}}, April 2023.

\bibitem[Vehtari et~al.(2017)Vehtari, Gelman, and
  Gabry]{2017vehtariPracticalBayesianModel}
Aki Vehtari, Andrew Gelman, and Jonah Gabry.
\newblock Practical {{Bayesian}} model evaluation using leave-one-out
  cross-validation and {{WAIC}}.
\newblock \emph{Statistics and Computing}, 27\penalty0 (5):\penalty0
  1413--1432, September 2017.
\newblock ISSN 1573-1375.
\newblock \doi{10.1007/s11222-016-9696-4}.

\bibitem[Bingham et~al.(2019)Bingham, Chen, Jankowiak, Obermeyer, Pradhan,
  Karaletsos, Singh, Szerlip, Horsfall, and
  Goodman]{2019binghamPyroDeepUniversal}
Eli Bingham, Jonathan~P. Chen, Martin Jankowiak, Fritz Obermeyer, Neeraj
  Pradhan, Theofanis Karaletsos, Rohit Singh, Paul Szerlip, Paul Horsfall, and
  Noah~D. Goodman.
\newblock Pyro: Deep universal probabilistic programming.
\newblock \emph{J. Mach. Learn. Res.}, 20\penalty0 (1):\penalty0 973--978,
  January 2019.
\newblock ISSN 1532-4435.

\bibitem[Silverman(1985)]{1985silvermanAspectsSplineSmoothing}
B.~W. Silverman.
\newblock Some {{Aspects}} of the {{Spline Smoothing Approach}} to
  {{Non-Parametric Regression Curve Fitting}}.
\newblock \emph{Journal of the Royal Statistical Society. Series B
  (Methodological)}, 47\penalty0 (1):\penalty0 1--52, 1985.
\newblock ISSN 0035-9246.

\bibitem[Yeafi(2021)]{2021yeafiPDBElectricPower}
Ashfak Yeafi.
\newblock {{PDB Electric Power Load History}}, 2021.

\bibitem[{Carreira-Perpi{\~n}{\'a}n} and
  Williams(2003)]{2003carreira-perpinanNumberModesGaussian}
Miguel~{\'A}. {Carreira-Perpi{\~n}{\'a}n} and Christopher K.~I. Williams.
\newblock On the {{Number}} of {{Modes}} of a {{Gaussian Mixture}}.
\newblock In Gerhard Goos, Juris Hartmanis, Jan Van~Leeuwen, Lewis~D. Griffin,
  and Martin Lillholm, editors, \emph{Scale {{Space Methods}} in {{Computer
  Vision}}}, volume 2695, pages 625--640, Isle of Skye, UK, June 2003. Springer
  Berlin Heidelberg.
\newblock ISBN 978-3-540-40368-5 978-3-540-44935-5.
\newblock \doi{10.1007/3-540-44935-3_44}.

\bibitem[Vehtari et~al.(2021)Vehtari, Gelman, Simpson, Carpenter, and
  B{\"u}rkner]{2021vehtariRankNormalizationFoldingLocalization}
Aki Vehtari, Andrew Gelman, Daniel Simpson, Bob Carpenter, and Paul-Christian
  B{\"u}rkner.
\newblock Rank-{{Normalization}}, {{Folding}}, and {{Localization}}: {{An
  Improved R{\textasciicircum}}} for {{Assessing Convergence}} of {{MCMC}}
  (with {{Discussion}}).
\newblock \emph{Bayesian Analysis}, 16\penalty0 (2):\penalty0 667--718, June
  2021.
\newblock ISSN 1936-0975, 1931-6690.
\newblock \doi{10.1214/20-BA1221}.

\bibitem[Wasserman(2004)]{2004wassermanAllStatisticsConcise}
Larry Wasserman.
\newblock \emph{All of {{Statistics}}: {{A Concise Course}} in {{Statistical
  Inference}}}.
\newblock Springer {{Texts}} in {{Statistics}}. Springer, New York, NY, 2004.
\newblock ISBN 978-1-4419-2322-6 978-0-387-21736-9.
\newblock \doi{10.1007/978-0-387-21736-9}.

\end{thebibliography}
	\newpage
	\appendix
	\section{Data and Code Licensing and Availability}
	Table~\ref{tab:assets} reports the digital assets used in the experiments, while our code is openly available through GitHub at {\url{https://github.com/aurelio-raffa/cocoafuse.git}}.
	\begin{table}[h!]
		\caption{Assets, Links and License Types}\label{tab:assets}
		\centering
		\begin{tabular}{lll}
			\toprule
			Asset & Website & License Type\\\midrule
			mcycle Dataset & \href{https://www.kaggle.com/datasets/nirmalsankalana/mcycle?resource=download}{Kaggle} & CC0: Public Domain\\
			mcycle annotations & \href{https://www.stat.cmu.edu/~larry/all-of-statistics/index.html}{Larry Wasserman's \textit{All of Statistics}} & \textemdash\\
			Wind Turbine Dataset & \href{https://www.kaggle.com/datasets/berkerisen/wind-turbine-scada-dataset?resource=download}{Kaggle} & \textemdash\\
			Power Distribution Board & \href{https://www.kaggle.com/datasets/ashfakyeafi/pbd-load-history}{Kaggle} & CC0: Public Domain\\
			Code & {\href{https://github.com/aurelio-raffa/cocoafuse.git}{GitHub}} & GNU GPLv3\\
			\bottomrule
		\end{tabular}
	\end{table}

	\section{Proofs}\label{app:proofs}
	\begin{proofs}[of Lemma~\ref{lemma:blend-is-a-density}]
		Equation~\eqref{eq:blend-definition} only contains (finitely many) non-negative terms, which means that $f_{\bm{\pi}}^{\textup{blend}}$ is non-negative.
		Now, observe that
		\begin{equation}
			 \int \frac{\sigma_i}{\sigmablend} f_i(x_i) \mathrm{d}x = \int f_i(x_i) \frac{\mathrm{d}x_i}{\mathrm{d}x} \mathrm{d}x = 1,
		\end{equation}
		hence $\int f_{\bm{\pi}}^{\textup{blend}}(x) \mathrm{d}x = \sum_{i=1}^M \pi_i = 1$.
	\end{proofs}
	\begin{proofs}[of Lemma~\ref{lemma:blend-of-gaussians}]
		Let the
		base densities $f_i = \mathcal{N}(\mu_i, \sigma_i)$ be non-degenerate (proper), i.e., $\sigma_i>0$ for all $i$;
		then, $\sigmablend^2 \ge \min_j\{\sigma_j^2\} > 0$.
		Further,
		\begin{align*}
             \frac{\sigma_i}{\sigmablend} f_i(x_i) &=
             \frac{\sigma_i}{\sigmablend \sqrt{2\pi \sigma_i^2}} \exp{\left\{-\frac{(x_i - \mu_i)^2}{\sigma_i^2}\right\}} =\\
            &=  \frac{1}{\sqrt{2\pi \sigmablend^2}} \exp{\left\{-\frac{\left(\sigma_i\frac{x - \mublend}{\sigmablend} + \mu_i - \mu_i\right)^2}{\sigma_i^2}\right\}} =\\
            &=  \frac{1}{\sqrt{2\pi \sigmablend^2}} \exp{\left\{-\frac{(x - \mublend)^2}{\sigmablend^2}\right\}} = f(x; \mublend, \sigmablend^2),
        \end{align*}
		therefore $f_{\bm{\pi}}^{\textup{blend}}(x) = \sum_{i=1}^M \pi_i f(x; \mublend, \sigmablend^2) = f(x; \mublend, \sigmablend^2) = \mathcal{N}(\mublend, \sigmablend)$.
	\end{proofs}
	\begin{proofs}[of Lemma~\ref{lemma:interpolation-is-a-blend}]
		By expanding on the expressions~\eqref{eq:blend-mus} and \eqref{eq:blend-sigmas},
		we can notice that~\eqref{eq:fusion-mus} and~\eqref{eq:fusion-sigmas} amount to a blend with weights $\bm{\tau}^{(i)}$ given by
		$\tau_j^{(i)} = \beta \delta_{ij} + (1 - \beta) \pi_j$,
		where $\delta_{ij}$ is the Kronecker delta.
	\end{proofs}
	\begin{proofs}[of Proposition~\ref{prop:number-of-modes}]
		We assume that the base densities are non-degenerate, i.e., $\sigma_i > 0$ for all $i$.
		Then, by Lemma~\ref{lemma:blend-of-gaussians}, each blend $f_i^{(\beta)}$ is non-degenerate.
		Further, we assume that $\pi_i \in (0, 1)$ for all $i$ and likewise $\beta \in (0, 1)$.
		Since the a fusion of $M$ proper Gaussian base densities is a mixture of $M$ proper Gaussians, the result holds by Theorem 2 of~\citet{2003carreira-perpinanNumberModesGaussian}.
	\end{proofs}
	\begin{proofs}[of Proposition~\ref{prop:switch-conditional-mean}]
		The conditional law of the target $y$ on the covariate $x$ can be expressed, upon introducing an auxiliary Bernoulli variable $Z$, as hierarchical model
		\begin{equation*}
            Z|x \sim \textup{Be}(p(x)),\quad Y|Z = z, x \sim \begin{cases}
                \mathcal{N}(\mu_1, \sigma)\quad\textup{if }z = 1,\\
                \mathcal{N}(\mu_2, \sigma)\quad\textup{if }z = 0,
            \end{cases}
        \end{equation*}
		so that the conditional density $p(y|x)$ can be computed, through marginalization of $Z$
		\begin{equation*}\label{eq:switch-conditional-law}
			 p(y|x) = \sum_{z=0}^1 p(y, Z = z | x) = \sum_{z=0}^1 p(y | Z = z, x) p(Z = z | x).
		\end{equation*}
		Therefore,
		\begin{equation*}
            \mathbb{E}\left[Y|X=x\right] = \int y p(y|x) \mathrm{d}y = \sum_{z=0}^1 p(Z = z | x) \int y p(y | Z = z, x) \mathrm{d}y = p(x)\mu_1 + \left(1-p(x)\right)\mu_2
        \end{equation*}
	\end{proofs}
	\begin{proofs}[of Proposition~\ref{prop:transition-conditional-mean}]
		In light of~\eqref{eq:transition-hierarchical}, it holds that
		\begin{equation}\label{eq:transition-conditional-law}
			p(y|x) =
			\int_{0}^1 p(y | \alpha, x) p(\alpha | x) d\alpha.
		\end{equation}
		By applying the Fubini-Tonelli theorem, we then obtain
		\begin{align}
			\nonumber     \mathbb{E}[Y | X = x] &= \int_{y\in \mathbb{R}} y p(y|x) dy = \int_{y\in \mathbb{R}} y \left(\int_{0}^1 p(y | \alpha, x) p(\alpha | x) d\alpha\right) dy =\\
			\nonumber	&= \int_{0}^1 (\mu_1\alpha + \mu_2(1 - \alpha))  p(\alpha | x) d\alpha =\\
			&= \mu_1 \mathbb{E}[\alpha | X = x] + \mu_2 \mathbb{E}[1 - \alpha | X = x] = p(x)\mu_1 + (1 - p(x))\mu_2.
		\end{align}
	\end{proofs}
	\begin{proofs}[of Proposition~\ref{prop:grad-entropy}]
		By expanding upon the definition~\eqref{eq:entropy},
		\begin{subequations}
			\begin{align}
				\nabla_{\bm{\lambda}} \mathcal{H}(\bm{\lambda}) &=\label{seq:25a}
				\int_{\Theta} \ln H\left(\bar{\bm{\alpha}}(\bm{\theta})\right) \nabla_{\bm{\lambda}} \textup{post}_{\bm{\lambda}}(\mathrm{d}\bm{\theta}) = \int_{\Theta} \ln H\left(\bar{\bm{\alpha}}(\bm{\theta})\right) \frac{\nabla_{\bm{\lambda}} \textup{post}_{\bm{\lambda}}(\bm{\theta})}{\textup{post}_{\bm{\lambda}}(\bm{\theta})} \textup{post}_{\bm{\lambda}}(\mathrm{d}\bm{\theta}) =\\
				&= \int_{\Theta} \ln H\left(\bar{\bm{\alpha}}(\bm{\theta})\right) \nabla_{\bm{\lambda}} \left( \ln\textup{post}_{\bm{\lambda}}(\bm{\theta}) \right)\textup{post}_{\bm{\lambda}}(\mathrm{d}\bm{\theta}) =\\
				&= \mathbb{E}_{\bm{\theta}\sim\textup{post}_{\bm{\lambda}}}\left[ \ln H\left(\bar{\bm{\alpha}}(\bm{\theta})\right) \nabla_{\bm{\lambda}} \left(\ln L(\bm{y} | \bm{X}, \bm{\theta}) + \ln p(\bm{\theta} | \bm{\lambda}) - \ln L(\bm{\lambda}; \bm{y}, \bm{X}) \right) \right] =\\
				&= \mathbb{E}_{\bm{\theta}\sim\textup{post}_{\bm{\lambda}}}\left[
				\ln H\left(\bar{\bm{\alpha}}(\bm{\theta})\right) \left( \nabla_{\bm{\lambda}} \ln p(\bm{\theta} | \bm{\lambda}) - \nabla_{\bm{\lambda}} \ln L(\bm{\lambda}; \bm{y}, \bm{X}) \right)
				\right] =\\
				&= \mathbb{E}_{\bm{\theta}\sim\textup{post}_{\bm{\lambda}}}\left[
				\ln H\left(\bar{\bm{\alpha}}(\bm{\theta})\right) \left( \nabla_{\bm{\lambda}} \ln p(\bm{\theta} | \bm{\lambda}) -  \mathbb{E}_{\bm{\theta}\sim\textup{post}_{\bm{\lambda}}}\left[ \nabla_{\bm{\lambda}} \ln p(\bm{\theta} | \bm{\lambda}) \right] \right)
				\right] =\\
				&= \begin{aligned}[t]\label{seq:25f}
					\mathbb{E}_{\bm{\theta}\sim\textup{post}_{\bm{\lambda}}}\left[
					\ln H\left(\bar{\bm{\alpha}}(\bm{\theta})\right) \nabla_{\bm{\lambda}} \ln p(\bm{\theta} | \bm{\lambda}) \right] - \mathbb{E}_{\bm{\theta}\sim\textup{post}_{\bm{\lambda}}}\left[ \ln H\left(\bar{\bm{\alpha}}(\bm{\theta})\right) \mathbb{E}_{\bm{\theta}\sim\textup{post}_{\bm{\lambda}}}\left[ \nabla_{\bm{\lambda}} \ln p(\bm{\theta} | \bm{\lambda}) \right]
					\right] =
				\end{aligned}\\
				&= \begin{aligned}[t]\label{seq:25g}
					\mathbb{E}_{\bm{\theta}\sim\textup{post}_{\bm{\lambda}}}\left[
					\ln H\left(\bar{\bm{\alpha}}(\bm{\theta})\right) \nabla_{\bm{\lambda}} \ln p(\bm{\theta} | \bm{\lambda}) \right] - \mathbb{E}_{\bm{\theta}\sim\textup{post}_{\bm{\lambda}}}\left[  \ln H\left(\bar{\bm{\alpha}}(\bm{\theta})\right) \right]\mathbb{E}_{\bm{\theta}\sim\textup{post}_{\bm{\lambda}}}\left[\nabla_{\bm{\lambda}} \ln p(\bm{\theta} | \bm{\lambda})
					\right] =
				\end{aligned}\\
				&= \begin{aligned}[t]\label{seq:25h}
					\mathbb{E}_{\bm{\theta}\sim\textup{post}_{\bm{\lambda}}}\left[
					\ln H\left(\bar{\bm{\alpha}}(\bm{\theta})\right) \nabla_{\bm{\lambda}} \ln p(\bm{\theta} | \bm{\lambda}) \right] - \mathbb{E}_{\bm{\theta}\sim\textup{post}_{\bm{\lambda}}}\left[ \mathbb{E}_{\bm{\theta}\sim\textup{post}_{\bm{\lambda}}}\left[  \ln H\left(\bar{\bm{\alpha}}(\bm{\theta})\right) \right]\nabla_{\bm{\lambda}} \ln p(\bm{\theta} | \bm{\lambda})
					\right] =
				\end{aligned}\\
				&=\label{seq:25j} \mathbb{E}_{\bm{\theta}\sim\textup{post}_{\bm{\lambda}}}\left[
				\left(\ln H\left(\bar{\bm{\alpha}}(\bm{\theta})\right) - \mathcal{H}(\bm{\lambda}) \right) \nabla_{\bm{\lambda}} \ln p(\bm{\theta} | \bm{\lambda})
				\right]
			\end{align}
		\end{subequations}
		where, in equation~\eqref{seq:25a}, $\textup{post}_{\bm{\lambda}}(\bm{\theta})$ denotes the posterior distribution of the model parameters $\bm{\theta}$ given priors indexed by $\bm{\lambda}$,
		and in equation~\eqref{seq:25f} we exploit the fact that $\mathbb{E}_{\bm{\theta}\sim\textup{post}_{\bm{\lambda}}}\left[ \nabla_{\bm{\lambda}} \ln p(\bm{\theta} | \bm{\lambda}) \right]$ is a constant with respect to the density $\textup{post}_{\bm{\lambda}}$, which allows us to factorize the expectation as in~\eqref{seq:25g} and rearrange the terms for convenience as in~\eqref{seq:25h}\footnote{This is merely for convenience of implementation.}.
	\end{proofs}

	\section{Numerically Robust Transformations of the Likelihood}\label{app:numerical-robustness}
	Because of the structure of the equations~\eqref{eq:moe-likelihood} and~\eqref{eq:cocoafuse-likelihood} describing the MoE's and CoCoAFusE's likelihoods, working in the log-space (as required by Stan) might entail numerical issues.
	Indeed, the logarithm of a sum of exponentials is not numerically robust, although it would appear in both models' log-likelihoods.
	In the following, we discuss how to circumvent this issue, providing numerically sound expressions for the MoE and CoCoAFusE log-likelihood densities, which are the ones ultimately implemented in our code\footnote{{\url{https://github.com/aurelio-raffa/cocoafuse.git}}}.

	\paragraph{MoE.}
	By taking the logarithm of equation~\eqref{eq:moe-likelihood}, we can express the log-likelihood as a function of the log-densities $\ln p(\bm{y} | \bm{x}; \bm{\theta}_i)$ and $\ln p(z = i |  \bm{x};\: \bm{\theta}_{\mathcal{G}})$, with $i\in\{1,\dots, M\}$, as
	\begin{align}\label{subeq:moe-log-likelihood}
		\nonumber	\ln p(\bm{y} | \bm{x}; \bm{\theta}) &= \ln \left(\sum_{i=1}^M p(\bm{y} | \bm{x}; \bm{\theta}_i) p(z = i |  \bm{x};\: \bm{\theta}_{\mathcal{G}})\right) =\\
		&= \ln \left( \sum_{i=1}^M \exp\left\{ \ln p(\bm{y} | \bm{x}; \bm{\theta}_i) + \ln p(z = i |  \bm{x};\: \bm{\theta}_{\mathcal{G}}) \right\} \right).
	\end{align}
	The last expression can be substituted with the more robust Log-Sum-Exp (LSE) $n$-ary function, already implemented in Stan.

	\paragraph{CoCoAFusE.}
	By taking the natural logarithm of both sides in~\eqref{eq:cocoafuse-likelihood}, we instead get to the following expression:
	\begin{align}\label{eq:cocoafuse-lse}
		\nonumber	\ln p(\bm{y} | \bm{x}; \bm{\theta}) &= \ln \left(\sum_{i=1}^M p(\bm{y} | \bm{x}; \bm{\theta}_i^{(\beta)}) p(z = i |  \bm{x};\: \bm{\theta}_{\mathcal{G}})\right) =\\
		&= \ln \left( \sum_{i=1}^M \exp\left\{ \ln p(\bm{y} | \bm{x}; \bm{\theta}_i^{(\beta)}) + \ln p(z = i |  \bm{x};\: \bm{\theta}_{\mathcal{G}}) \right\} \right).
	\end{align}

	\section{Choice of Prior Distributions}\label{app:choice-of-distr}

	\paragraph{Priors on the Expert Parameters}
	The Gaussian conditional distribution $p(y | \bm{x}; \bm{\theta}_i)$ of the target introduced in Remark~\ref{remark:gaussian-conditional-experts} as well as the categorical one for $p(\bm{z} | \bm{x}; \bm{\theta}_{\mathcal{G}})$ introduced in Remark~\ref{remark:gate-parameterization} are not the only possible choices of distributions. Nonetheless, such distributions are already flexible enough to accommodate the modeling scenarios presented in the work.
	In view of parameterization~\eqref{eq:gaussian-expert-moe},
	we simply need to assign priors on the $\bm{\theta}_i$ and $\sigma_i$ variables, which is also compatible with~\eqref{eq:cocoa-fused-experts}.
	We choose Laplace-distributed independent priors on the $\bm{\theta}_i$, with the defaults for the location and scale parameters being 0 and 1, respectively.
	To complete the set of priors on the expert, we assign apriori independent log-normal densities on the standard errors $\sigma_i$.

	\paragraph{Priors on the Gate Parameters}
	We choose Laplace-distributed independent priors on the gate coefficients, with the defaults for the location and scale parameters being 0 and 1, respectively.
	We choose Laplace distributions in light of the slower decay of the tails compared to the Gaussian distribution, which gives more flexibility to the model.
	It is worth remarking that in our implementation, this prior is elected on all but the elements of the last column of the matrix $\bm{\theta}_\mathcal{G}$, in light of the additional observations made in Section~\ref{ssec:label-switching-problem}. In this way, the gate's parameters are well-defined and can be identified. As a last note, when one considers $M = 2$ experts, $\bm{\theta}_\mathcal{G}^\top\Phi(\bm{x})$ simply reduces to the logit of the probability of selecting the first expert.

\section{Emprical Bayes Approach for Prior Elicitation}\label{app:empirical-bayes}

Propositions~\ref{prop:witten-datamining} and~\ref{prop:grad-entropy} justify
a gradient ascent scheme on $\bm{\lambda}$ for the loss function~\eqref{eq:eb-loss}.
Consider following the Monte Carlo estimates for $\textup{ELL}(\bm{\lambda})$ and the gradients of the log-marginal likelihood:
\begin{equation}\label{eq:mc-approx-ell}
	 \textup{ELL}(\bm{\lambda})\approx \widehat{\textup{ELL}}_S(\bm{\lambda}) =\frac{1}{S} \sum_{s=1}^{S}\sum_{n=1}^{N} \ln p\left(\bm{y}_n \Big\vert \bm{x}_n; \bm{\theta}^{(s)}\right) + \frac{1}{S}\sum_{s=1}^{S} \ln  p\left(\bm{\theta}^{(s)} \Big\vert \bm{\lambda}\right),
\end{equation}
\begin{equation}\label{eq:mc-approx-grads}
	 \mathbb{E}_{\bm{\theta}\sim\textup{post}_{\bm{\lambda}}}\left[ \nabla_{\bm{\lambda}} \ln L(\bm{\lambda};\bm{\theta}, \bm{X}, \bm{y}) \right] \approx \frac{1}{S}\sum_{s=1}^{S} \nabla_{\bm{\lambda}} \ln  p\left(\bm{\theta}^{(s)} \Big\vert \bm{\lambda}\right),
\end{equation}
with $\left\{\bm{\theta}^{(s)}\right\}_{s=1}^S$ being a sample from the posterior.
Notice that quantities in formulas~\eqref{eq:mc-approx-ell} and~\eqref{eq:mc-approx-grads} can be easily computed analytically.
Likewise, the evaluations and gradients of~\eqref{eq:entropy} can be approximated via Proposition~\ref{prop:grad-entropy} via
\begin{equation}\label{eq:log-entropy-surrogate}
	 \mathcal{H}(\bm{\lambda}) \approx \widehat{\mathcal{H}}_S(\bm{\lambda}) = \frac{1}{S}\sum_{s=1}^S \ln H\left(\frac{1}{N}\sum_{n=1}^N\bm{\alpha}\left(\bm{x}_n; \bm{\theta}^{(s)}\right)\right),
\end{equation}
\begin{equation}\label{eq:log-entropy-grads}
	 \nabla_{\bm{\lambda}} \mathcal{H}(\bm{\lambda}) \approx \frac{1}{S} \sum_{s=1}^{S} \left(\ln H\left(\frac{1}{N}\sum_{n=1}^N\bm{\alpha}\left(\bm{x}_n; \bm{\theta}^{(s)}\right)\right) - \widehat{\mathcal{H}}_S(\bm{\lambda})\right)\nabla_{\bm{\lambda}} \ln  p\left(\bm{\theta}^{(s)} \Big\vert \bm{\lambda}\right).
\end{equation}
While we can estimate the gradients of $\mathcal{L}(\bm{\lambda})$ via~\eqref{eq:mc-approx-grads} and~\eqref{eq:log-entropy-grads},
we evaluate individual hyperparameter iterates $\bm{\lambda}^{(k)}$ via the surrogate cost function
\begin{equation}
	\hat{\mathcal{L}}\left(\bm{\lambda}^{(k)}\right) = \widehat{\textup{ELL}}_S\left(\bm{\lambda}^{(k)}\right) + \gamma \widehat{\mathcal{H}}_S\left(\bm{\lambda}^{(k)}\right).
\end{equation}
Since all of the evaluations are stochastic, we pick the best iterate over a fixed horizon.
	\section{Model selection}\label{app:model-select}
	\begin{algorithm}[b!]
		\caption{Complexity-Aware Model Selection Procedure}\label{alg:pareto-model-selection}
		\begin{algorithmic}
			\Require \textit{metric}$()$ (performance index functional)
			\Require \textit{complexity.nExperts}$()$ (number of experts)
			\Require \textit{complexity.nParams}$()$ (number of parameters)
			\Require $\tau$ (probability lower bound threshold for selection)
			\Require trials (experiments on the model with different hyperparameters)
			\State $\textup{paretoSet} \gets \textup{pareto}(\textup{trials};\ \textup{\textit{metric}},\ \textup{\textit{complexity.nExperts}},\ \textup{\textit{complexity.nParams}})$
			\State $\textup{bestTrial} \gets \varnothing$
			\For{$p \textup{ \textbf{in} } \textup{sortAscending}(\textup{paretoSet};\ \textup{\textit{complexity.nExperts}},\ \textup{\textit{complexity.nParams}})$}
			\If{$\textup{bestTrial} = \varnothing$ \textbf{or} $\textup{ChebyshevLB}(\textup{\textit{metric}}(\textup{bestTrial}) < \textup{\textit{metric}}(\textup{p})) > \tau$}
			$\textup{bestTrial} \gets p$
			\EndIf
			\EndFor\\
			\Return bestTrial
		\end{algorithmic}
	\end{algorithm}
	After having trained multiple models for different hyperparameter guesses obtained, e.g., via grid-search, we adopt PSIS-LOO as a conventional performance indicator guiding our choice of the model structure.
	Subsequently, the Pareto frontier of the resulting configurations is explored for increasing number of experts and complexity (in this order). For every new trial we visit, we compute the Chebyshev lower bound on the probability that the current iterate has a higher metric than the previous best. If the Chebyshev lower bound exceeds a predetermined threshold $\tau \in (0, 1)$, then we accept the current trial as best. Otherwise, we discard it even when the computed metric is higher until termination.%\\
	The previous is summarized as pseudo-code in Algorithm~\ref{alg:pareto-model-selection}.

	\paragraph{Remarks} First of all, the \emph{order} in which we visit the trials is relevant and should not be changed. Indeed, it reflects our priority of choosing models with as few experts as possible, breaking ties by favoring the least complex models (i.e., involving fewer uncertain parameters). Second, in our implementation the Chebyshev lower bound computation is carried out assuming the two model evaluations are independent, and each has mean and standard deviation equal to the computed PSIS-LOO score and PSIS-LOO standard error. Under these assumptions, the Chebyshev lower bound is a \textit{pessimistic} estimate of the probability of improvement under more complex configurations. Hence, one can trade (expected) performance improvement with model complexity increase by adjusting the value of $\tau$. In all our tests, we employ a fixed threshold $\tau = 0.5$. This choice implies that we want to be convinced that, under our assumptions, we improve performance more often than not. Last but not least, this procedure is not exact and does not fully guarantee the exclusion of pathological cases relating to expert collapse. Nonetheless, we empirically found that it helps to select only those trials where additional complexity is justified by substantial improvements in model performance.

	\section{Details on the Numerical Examples}\label{app:numerical-examples}

	\paragraph{General constructive details.}
	In both cases, the training set comprises $500$ observations, while $500$ samples are employed for testing.
	We perform hyperparameter tuning via the procedure described in Appendix~\ref{app:model-select} on the number of experts (with $M\in \{2, 3\}$) as well as the degree of the polynomial features for the gate, paying special attention to whether the correct number of experts ($M = 2$ in both examples) is identified to approximate the conditional density $p(y|x)$.

	\paragraph{Additional results on the ``Transition'' case.}
	While the CoCoAFusE and MoE with $M = 3$ experts do not exhibit large gaps in performance (see Table~\ref{tab:transition-scores}),
	CoCoAFusE performs better in terms of Log-Pointwise Predictive Density (LPPD) and expected Mean Squared Error (eMSE) on the test set.
	For completeness in \tablename{~\ref{tab:transition-scores}} and \figurename{~\ref{fig:transition-posterior-predictive}} are included the results attained by training a MoE with the correct number of experts.
	As shown there, setting $M=2$ leads to poor performance of the parsimonious MoE, indicating that one cannot reproduce the soft transition behavior with such a model. Meanwhile, these results show that the complexity overhead introduced by CoCoAFusE does not rest on the experts but solely on the behavior gate module.

	\section{Details on the Benchmark Examples}\label{app:details-benchmarks}
	\subsection{Convergence Diagnostics}\label{app:convergence-diagnostics}
		For models that were fitted via MCMC (i.e., MoE, BoE, CoCoAFusE, and BNN) we report the rank-normalized $\hat{R}$ Score (see~\citet{2021vehtariRankNormalizationFoldingLocalization}) in Table~\ref{tab:convergence-diagnostics}.
		Notice that the BNN consistently exceeds the conventional threshold of 1.05, indicating the possibility of a failed convergence.

	\begin{table}[b!]
		\centering
		\caption{$\hat{R}$ Statistic (should be less than 1.05, violations in \textbf{bold}) by Experiment and Model}\label{tab:convergence-diagnostics}
		\begin{tabular}{lcccc}\toprule
			Example & MoE & BoE & CoCoAFusE & BNN \\\midrule
			Wind Turbine & 1.013 & 1.036 & 1.023 & \textbf{1.245} \\
			Motorcycle Crash & 1.011 & 1.009 & 1.010 & \textbf{1.257} \\
			Power Distribution Board & 1.007 & 1.010 & 1.019 & \textbf{2.762} \\\bottomrule
		\end{tabular}
	\end{table}

	\subsection{Further details on the Wind Turbine example}\label{app:wind-turbine}
	The available data are gathered in the Yalova wind farm, which comprises 6 wind turbines with a combined nominal power of 54000 kW and employs a SCADA system to record turbine data, such as wind speed, wind direction, and generated power at 10-minute intervals. The original data also supplies the theoretical power curve values (in kWh) provided by the turbine manufacturer corresponding to the measured wind speed at the hub height.
	The dataset contains upward of 50000 time-stamped observations of four variables:
	\begin{description}[nosep]
		\item[\texttt{active\_power},] i.e., the output power generated at the recorded time, target of our analysis;
		\item[\texttt{wind\_speed},] i.e., the total speed of the wind;
		\item[\texttt{theoretical\_pc},] i.e., the nominal value of generation at the observed wind speed, as reported by the turbine manufacturer;
		\item[\texttt{wind\_direction},] i.e., the direction (in degrees) of the wind with respect to the alignment of the turbine. Note that, turbines automatically align to the wind direction.
	\end{description}
	We normalize the wind speeds by the maximum absolute value in the training set. Meanwhile, the theoretical curve and the actual power are normalized by their largest theoretical value contained in the training set, corresponding to a plateau in the theoretical curve.
	Upon inspection of the aforementioned, the following \textit{regimes} emerge.
	\begin{enumerate}[nosep]
		\item An initial plateau where both the active power and theoretical curve are zero, corresponding to wind speeds falling below the cut-in threshold.
		\item A final saturation plateau where both the active power and theoretical curve reach the maximum theoretical value for the turbine, with a mild noise affecting active power.
		\item A large intermediate region, where the difference between the theoretical curve and the actual power is noticeable and has a significant variance.
		\item An ``anomalous'' regime\footnote{We consider these points ``anomalous'' for lack of better explanation in the dataset description page.}, where the measured power is zero regardless of wind speeds.
	\end{enumerate}
	In order to treat the wind direction in accordance with its periodicity (i.e., 0$^\circ$ and 360$^\circ$ directions yielding the same conditional effects given the other covariates), we map the \texttt{wind\_direction} scalar to the sine and cosine of the corresponding angle.

	\paragraph{Priors.} The available Wind Turbine data are used to construct both an MoE, BoE, and a CoCoAFusE with $M=2$ experts (see Section~\ref{sec:experiments}), which we introduce based on our priors on the various operating modes of the system. In particular,
	the first expert encodes null power generation, while the second expert corresponds to theoretical power curve of the turbine.
	Priors are identical for MoE, BoE, and CoCoAFusE and are reported in~\tablename{~\ref{tab:hps-wind-turbine}}. Parameter blocks not listed in the table are given the default values as discussed in Appendix~\ref{app:choice-of-distr}.

	\begin{table}
		\centering
		\caption{Prior hyperparameters (location; scale) for the Wind Turbine Example.}\label{tab:hps-wind-turbine}
		\small
		\begin{tabular}{llrlrlrl}
			\toprule
			%			& & \multicolumn{3}{c}{Expert}\\\cmidrule{3-5}
			Parameter Block & Feature & \multicolumn{2}{c}{Exp. 1} & \multicolumn{2}{c}{Exp. 2} & \multicolumn{2}{c}{$\beta$} \\\midrule
			Gate& bias & (0.0; & 1.0) & (0.0; & 1.0) & \multicolumn{2}{c}{\textemdash} \\
			& \texttt{x.wind\_speed} & (0.0; & 1.0) & (0.0; & 1.0) & \multicolumn{2}{c}{\textemdash} \\
			& \texttt{x.sin\_wind\_direction} & (0.0; & 0.2) & (0.0; & 0.2) & \multicolumn{2}{c}{\textemdash} \\
			& \texttt{x.cos\_wind\_direction} & (0.0; & 0.2) & (0.0; & 0.2) & \multicolumn{2}{c}{\textemdash} \\\midrule
			Behaviour & bias & \multicolumn{2}{c}{\textemdash} & \multicolumn{2}{c}{\textemdash} & (5.0; & 0.15) \\
			& \texttt{x.wind\_speed} & \multicolumn{2}{c}{\textemdash} & \multicolumn{2}{c}{\textemdash} & (-10.0; & 0.15)  \\
			& \texttt{x.sin\_wind\_direction} & \multicolumn{2}{c}{\textemdash} & \multicolumn{2}{c}{\textemdash} & (0.0; & 0.2) \\
			& \texttt{x.cos\_wind\_direction} & \multicolumn{2}{c}{\textemdash} & \multicolumn{2}{c}{\textemdash} & (0.0; & 0.2)\\\midrule
			Experts & bias & (0.0; & 0.2) & (0.0; & 0.2)& \multicolumn{2}{c}{\textemdash} \\
			& \texttt{x.theoretical\_pc} & (0.0; & 0.07) & (1.0; & 0.07) & \multicolumn{2}{c}{\textemdash} \\\midrule
			Variances & \textemdash & (-4.0; & 0.2) & (-2.5; & 0.2) & \multicolumn{2}{c}{\textemdash} \\
			\bottomrule
		\end{tabular}
	\end{table}

	\paragraph{Interpretation of the results.} For datasets such as the one considered in this example, i.e., characterized by possible anomalies, having both tight predictive bounds and multiple overlapping descriptions is very important, as one can rely on the model to detect potential spurious observations at test time. Indeed, the CoCoAFusE's predictive density exhibits a bimodality between the two saturation plateaux, which matches the zero-active power observations. In this range, the probability of drawing from the corresponding expert can thus be used as an estimate for the anomalous state (e.g., a \textit{failure mode} for the physical object), further distinguishing it from
	the points laying in extremely low-density regions representing outliers or otherwise unmodeled anomalies.
	\begin{table}[tb!]
		\caption{Figures of Merit (Best in bold \textpm\ Std. Error) on the Switch Example.}
		\label{tab:switch-scores}
		\centering
		\begin{tabular}{lccccc}
			% table generated automatically on date 2025-01-31
			\toprule
			& Train & \multicolumn{4}{c}{Test} \\

			\cmidrule(lr){2-2}\cmidrule(lr){3-6}
			& $\uparrow$ LOO
			& $\uparrow$ LPPD
			& $_\uparrow^\downarrow$ 95\% CIC
			& $\downarrow$ 95\% CIL
			& $\downarrow$ eMSE \\

			\midrule

			% data extracted automatically on date 2025-01-31 from path outputs/switch/moe/inference/evaluation.csv
			MoE &
			-96.75\textpm18.10 &
			\textbf{-100.51}\textpm18.63 &
			\textbf{0.94}\textpm0.01 &
			3.15\textpm0.12 &
			3.98\textpm0.35\\

			% data extracted automatically on date 2025-01-31 from path outputs/switch/boe/inference/evaluation.csv
			BoE &
			-646.62\textpm41.13 &
			-680.61\textpm41.78 &
			0.89\textpm0.01 &
			3.33\textpm0.08 &
			\textbf{3.05}\textpm0.26\\

			% data extracted automatically on date 2025-01-31 from path outputs/switch/cocoafuse/inference/evaluation.csv
			CoCoAFusE &
			\textbf{-97.72}\textpm18.15 &
			-101.73\textpm18.65 &
			\textbf{0.94}\textpm0.01 &
			\textbf{3.11}\textpm0.12 &
			3.97\textpm0.35\\

			\bottomrule
		\end{tabular}
	\end{table}
	\begin{table}[tb!]
		\caption{Figures of Merit (Best in bold \textpm\ Std. Error) on the Transition Example.}
		\label{tab:transition-scores}
		\centering
		\begin{tabular}{lccccc}
			% table generated automatically on date 2025-01-31
			\toprule
			& Train & \multicolumn{4}{c}{Test} \\

			\cmidrule(lr){2-2}\cmidrule(lr){3-6}
			& $\uparrow$ LOO
			& $\uparrow$ LPPD
			& $_\uparrow^\downarrow$ 95\% CIC
			& $\downarrow$ 95\% CIL
			& $\downarrow$ eMSE \\

			\midrule

			% data extracted automatically on date 2025-01-31 from path outputs/transition/moe/inference/evaluation.csv
			MoE\textsubscript{3} &
			-144.99\textpm19.53 &
			-112.49\textpm14.45 &
			0.97\textpm0.01 &
			1.40\textpm0.02 &
			0.24\textpm0.01\\

			% data extracted automatically on date 2025-01-31 from path outputs/transition/moe2ex/inference/evaluation.csv
			MoE\textsubscript{2} &
			-400.22\textpm21.95 &
			-415.83\textpm21.86 &
			0.98\textpm0.01 &
			2.12\textpm0.04 &
			0.58\textpm0.03\\

			% data extracted automatically on date 2025-01-31 from path outputs/transition/boe/inference/evaluation.csv
			BoE &
			-141.10\textpm22.15 &
			-104.30\textpm15.37 &
			\textbf{0.96}\textpm0.01 &
			\textbf{1.21}\textpm0.00 &
			\textbf{0.19}\textpm0.01\\

			% data extracted automatically on date 2025-01-31 from path outputs/transition/cocoafuse/inference/evaluation.csv
			CoCoAFusE &
			\textbf{-113.61}\textpm17.70 &
			\textbf{-97.81}\textpm14.71 &
			\textbf{0.96}\textpm0.01 &
			1.22\textpm0.01 &
			\textbf{0.19}\textpm0.01\\

			\bottomrule
		\end{tabular}
	\end{table}
	\begin{figure}[tb!]
		\centering% .495
		\subfigure[MoE, $M = 3$ (Algorithm~\ref{alg:pareto-model-selection})]{\label{subfig:transition-moe3ex}\includegraphics[width=0.328\linewidth, trim={0.25cm 0 0.45cm 0.25cm}, clip]{Figures/transition/moe/y_vs_x.pdf}}
		\subfigure[MoE, $M = 2$ (Fixed)]{\label{subfig:transition-moe2ex}\includegraphics[width=0.328\linewidth, trim={0.25cm 0 0.45cm 0.25cm}, clip]{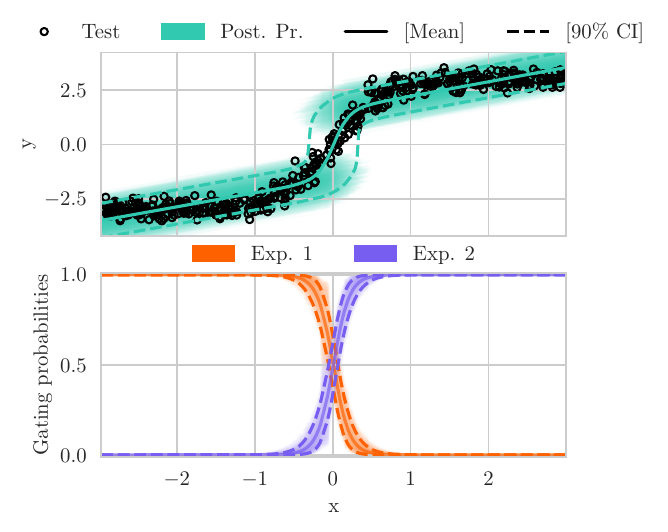}}
		\subfigure[CoCoA., $M = 2$ (Algorithm~\ref{alg:pareto-model-selection})]{\label{subfig:transition-cocoa}\includegraphics[width=0.328\linewidth, trim={0.25cm 0 0.45cm 0.25cm}, clip]{Figures/transition/cocoafuse/y_vs_x.pdf}}

		\caption{Predictions for MoEs and CoCoAFusE on the Transition Example.}
		\label{fig:transition-posterior-predictive2}
	\end{figure}

	\subsection{Further details on the Motorcycle Crash example}\label{app:motorcycle-crash}
	The Motorcycle Crash dataset consists of a series of accelerometer measurements for a simulated motorcycle crash experiment, which was used to test the efficacy of helmets. This dataset contains very few samples, i.e., 133 data points as reported in Section~\ref{sec:experiments}. Of these samples, 94 unique-timestamped points have been annotated in~\citet{2004wassermanAllStatisticsConcise} with an estimate of the different \enquote{regimes} to which the data belongs and the within-group residual variance.
	For what concerns the data considered for training and testing, we use as testing set both the 39 samples discarded by~\citeauthor{2004wassermanAllStatisticsConcise} and an additional 20\% of samples selected at random from the 94 unique-time observations, resulting in a training set of 75 observations and a test set of 58.
	In this case, we normalize the time coordinate (the only original covariate) for it to be mapped to the interval $[0, 1]$. Meanwhile, we divide the acceleration, i.e., the response, by its maximum absolute value within the training set to obtain training observations within $[-1, 1]$.
	Since we know that oscillations are present in the dataset, we introduce two additional features, defined as the sine and cosine of $\frac{2\pi t}{0.4}$ (where $t$ is the normalized time coordinates); in this way, all sinusoids of period 0.4 can be obtained as linear combinations of the newly defined features.
	We also define a quadratic feature for the normalized time, defined as $q(t) = \frac{64}{9} t (\frac{3}{4} - t)$, chosen as to have null correlation\footnote{Correlations among features can harm convergence of the HMC.} when $t\sim \textup{Unif}[0, 1]$.

	\paragraph{Priors.}
	Although our models do not rely on explicit group labels, this partition allows us to fix $M = 3$ and to select informed priors to attain tighter posterior predictive bounds.
	The full overview of prior hyperparameters for this example is reported in~\tablename{~\ref{tab:hps-motorcycle}}.
	\begin{table}[tb!]
		\centering
		\caption{Prior hyperparameters (location; scale) for the Motorcycle Crash Example.}\label{tab:hps-motorcycle}
		\small
		\begin{tabular}{llrlrlrl}
			\toprule
			Parameter Block & Feature & \multicolumn{2}{c}{Exp. 1} & \multicolumn{2}{c}{Exp. 2} & \multicolumn{2}{c}{Exp. 3} \\\midrule
			Gate
			& bias & (15.0; & 1.0) & (-50.0; & 1.0) & \multicolumn{2}{c}{\textemdash}\\
			& \texttt{x.times} & (-30.0; & 0.1)  & (50.0; & 0.1) & \multicolumn{2}{c}{\textemdash}\\
			& \texttt{x.times\_quad} & (0.0; & 10\textsuperscript{-2}) & (0.0; & 10\textsuperscript{-2}) & \multicolumn{2}{c}{\textemdash}\\\midrule
			Experts
			& bias & (0.0; & 1.0) & (0.0; & 1.0) & (0.0; & 0.5) \\
			& \texttt{x.cos\_times} & (0.0; & 10\textsuperscript{-2}) & (0.0; & 10\textsuperscript{-2}) & (0.0; & 2.0)\\
			& \texttt{x.sin\_times} & (0.0; & 10\textsuperscript{-2}) & (0.0; & 10\textsuperscript{-2}) & (0.0; & 2.0)\\\midrule
			Variances & \textemdash & (-5.0; & 0.2) & (-3.0; & 0.2) & (-1.0; & 0.2) \\
			\bottomrule
		\end{tabular}
	\end{table}

	\subsection{Further details on the Power Distribution Board example}

	In order to guarantee periodicity of the predictions over courses of days, weeks, and months, we transform the time coordinates (i.e., hour, weekday, month) into sine/cosine transformations with frequencies that are multiple of the ``natural'' one of the coordinate (i.e., $2\pi/24$ hours\textsuperscript{-1}, $2\pi/7$ days\textsuperscript{-1}, $2\pi/12$ months\textsuperscript{-1}), akin to a truncated Fourier series.
	We split the low- (month, week) and high-frequency features (day, hour) and map them to the gate and expert, respectively.
	This design choice reflects a supposed preference of the end user to achieve the highest interpretability on the prediction for a specified \textit{day}, with the fusion of experts directly encoding the hourly profile.
	Further, the temperature information is provided to both gating and experts, with the gating also receiving a quadratic transformation of the temperature to capture expert transitions that happen far from extreme values.

	\paragraph{Priors.}
	In this example, we rely on the aforementioned default initialization for the prior hyperparameters and perform the procedure described in Appendix~\ref{app:empirical-bayes} to tune the priors on the CoCoAFusE model. Identical priors (with the exception of the additional gate $\beta$) are then imposed on the MoE and BoE.

\end{document}